%% file: example_paper.tex
\documentclass{article}
\usepackage{times}

\usepackage{microtype}
\usepackage{graphicx}
\usepackage{subcaption}
\usepackage{booktabs}
\usepackage{makecell}
\usepackage{xcolor}
\usepackage{threeparttable}
\usepackage{caption}
\usepackage{enumitem}
\usepackage{multirow}
\usepackage{xspace}
\usepackage{wrapfig}
\usepackage{pifont}
\usepackage{tcolorbox}
\usepackage{algorithmic}
\usepackage{algorithm}
\usepackage{fancyhdr}

\usepackage{amsmath}
\usepackage{amssymb}
\usepackage{mathtools}
\usepackage{amsthm}

\usepackage[round]{natbib}
\setcitestyle{authoryear,round,citesep={;},aysep={,},yysep={;}}
\renewcommand{\cite}[1]{\citep{#1}}

\usepackage{listings}
\newcommand{\model}{\text{MAS4TS}\xspace}

\newcommand{\boldres}[1]{{\textbf{\textcolor{red}{#1}}}}
\newcommand{\secondres}[1]{{\underline{\textcolor{blue}{#1}}}}

\usepackage{hyperref}
\definecolor{mydarkblue}{rgb}{0,0.08,0.45}
\hypersetup{
    colorlinks=true,
    linkcolor=mydarkblue,
    citecolor=mydarkblue,
    filecolor=mydarkblue,
    urlcolor=mydarkblue
}

\newcommand{\RETURN}{\STATE \textbf{return} }

\usepackage[capitalize,noabbrev]{cleveref}

\theoremstyle{plain}

\theoremstyle{definition}

\theoremstyle{remark}

\makeatletter
\def\@normalsize{\@setsize\normalsize{11pt}\xpt\@xpt}
\def\small{\@setsize\small{10pt}\ixpt\@ixpt}
\def\footnotesize{\@setsize\footnotesize{10pt}\ixpt\@ixpt}
\def\scriptsize{\@setsize\scriptsize{8pt}\viipt\@viipt}
\def\tiny{\@setsize\tiny{7pt}\vipt\@vipt}
\def\large{\@setsize\large{14pt}\xiipt\@xiipt}
\def\Large{\@setsize\Large{16pt}\xivpt\@xivpt}
\def\LARGE{\@setsize\LARGE{20pt}\xviipt\@xviipt}
\def\huge{\@setsize\huge{23pt}\xxpt\@xxpt}
\def\Huge{\@setsize\Huge{28pt}\xxvpt\@xxvpt}
\makeatother

\makeatletter
\renewcommand\section{\@startsection{section}{1}{\z@}{-0.12in}{0.02in}{\large\bf\raggedright}}
\renewcommand\subsection{\@startsection{subsection}{2}{\z@}{-0.10in}{0.01in}{\normalsize\bf\raggedright}}
\renewcommand\subsubsection{\@startsection{subsubsection}{3}{\z@}{-0.08in}{0.01in}{\normalsize\sc\raggedright}}
\renewcommand\paragraph{\@startsection{paragraph}{4}{\z@}{1.5ex plus 0.5ex minus .2ex}{-1em}{\normalsize\bf}}
\makeatother

\def\toptitlebar{\hrule height1pt \vskip .25in}
\def\bottomtitlebar{\vskip .22in \hrule height1pt \vskip .3in}

\renewenvironment{abstract}
{\centerline{\large\bf Abstract}\vspace{-0.12in}\begin{quote}}
{\par\end{quote}\vskip 0.12in}

\renewcommand{\headrulewidth}{0pt}
\begin{document}
\twocolumn[
  \begin{center}
    \toptitlebar
    {\Large\bf Visual Reasoning over Time Series via Multi-Agent Systems}
    \bottomtitlebar
    {\normalsize
    \begin{tabular}{c}
      Weilin Ruan\textsuperscript{1} \quad
      Yuxuan Liang\textsuperscript{*,1} \\[0.5em]
      \textsuperscript{1}The Hong Kong University of Science and Technology (Guangzhou), China \\[0.3em]
      \texttt{rwlinno@gmail.com, yuxliang@outlook.com}
    \end{tabular}
    }
  \end{center}  
%  \vskip 0.3in
]

% Affiliations footnote
% {\let\thefootnote\relax\footnotetext{\hspace*{-\footnotesep}\textsuperscript{*} Corresponding author.

% \textsuperscript{1}Department of XXX, University of YYY. \textsuperscript{2}Company Name. \textsuperscript{3}School of ZZZ, Institute of WWW.
%
% }}

\thispagestyle{plain}

\input{contents/00abstract}

\input{contents/01introduction}
\input{contents/02relatedwork}
\input{contents/03method}

\input{contents/04experiments}
\input{contents/05conclusion}

\small
\bibliographystyle{abbrvnat}
\bibliography{example_paper}

\newpage
\onecolumn
\appendix
\input{appendix/A_Experimental_Details}
\input{appendix/B_Complete_Results}
\input{appendix/tool_library}
\input{appendix/prompts}
\input{appendix/justification}
\input{appendix/limitation}

\end{document}

%% file: contents/00abstract.tex
\begin{abstract}
Time series analysis underpins many real-world applications, yet existing time-series-specific methods and pretrained large-model-based approaches remain limited in integrating intuitive visual reasoning and generalizing across tasks with adaptive tool usage.
To address these limitations, we propose \textbf{\model}, a tool-driven multi-agent system for general time series tasks, built upon an \textbf{Analyzer--Reasoner--Executor} paradigm that integrates agent communication, visual reasoning, and latent reconstruction within a unified framework.
\model first performs visual reasoning over time series plots with structured priors using a Vision-Language Model to extract temporal structures, and subsequently reconstructs predictive trajectories in latent space.
Three specialized agents coordinate via shared memory and gated communication, while a router selects task-specific tool chains for execution.
Extensive experiments on multiple benchmarks demonstrate that \model achieves state-of-the-art performance across a wide range of time series tasks, while exhibiting strong generalization and efficient inference.
%Code is available at \url{https://github.com/anonymous/MAS4TS}.
\end{abstract}

%% file: contents/01introduction.tex
\begin{figure}[t!]
    \centering    
    \includegraphics[width=0.48\textwidth]{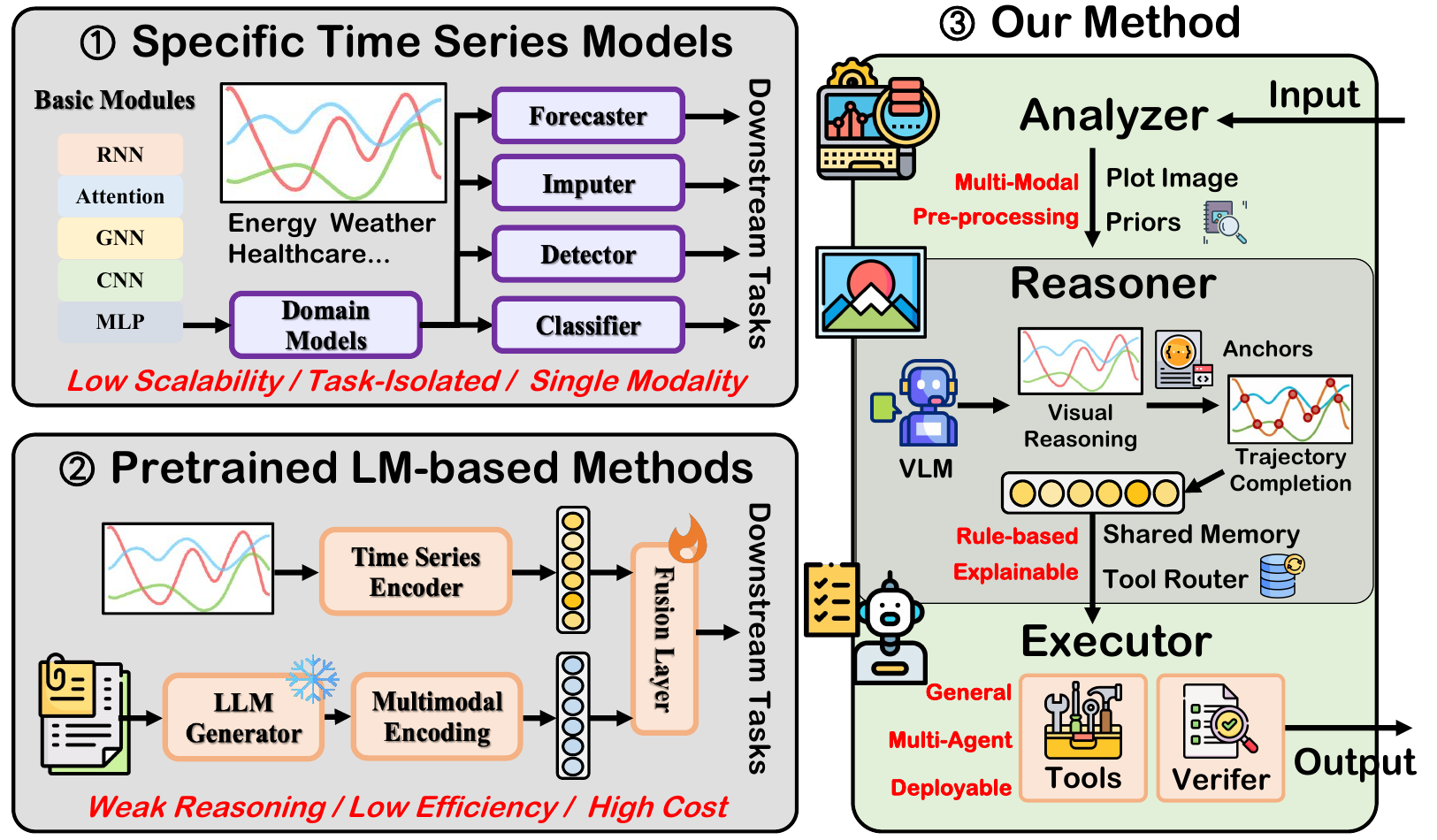}
    \caption{The motivation of \model. Existing approaches (left) struggle to support intuitive reasoning and task-adaptive execution, while \model (right) organizes time series analysis into a unified paradigm that collaboratively analyzes, reasons, and executes.}
    \label{fig:motivation}
\vspace{-4mm}
\end{figure}

\section{Introduction}

Time series analysis is fundamental to a wide range of real-world applications, including energy systems~\cite{deb2017review}, transportation~\cite{zheng2020traffic}, and other critical domains~\cite{idrees2019prediction,karevan2020transductive,schneider1974climate}.
Core tasks such as forecasting, classification, imputation, and anomaly detection~\cite{lim2021time} play a crucial role in supporting data-driven decision-making in these scenarios.

Recent advances in pretrained Large Language Models (LLMs) have demonstrated strong capabilities in pattern understanding and general reasoning~\cite{brown2020language,touvron2023llama}.
Motivated by this progress, a growing body of work explores leveraging pretrained models for time series analysis~\cite{jin2023time,liu2024unitime,zhou2023one}, aiming to improve generalization and contextual modeling beyond traditional neural architectures~\cite{wu2023timesnet,zhou2021informer,nie2022time}.
In parallel, agentic methods and multi-agent systems (MAS)~\cite{ravuru2024agentic,gu2025argos,chang2025survey,wu2024autogen,hong2023metagpt,qian2023chatdev,huang2025many} are emerging as a practical paradigm for tool orchestration, task decomposition, and coordinated execution.

Despite these advances, the application of multi-agent paradigms to general time series analysis remains largely unexplored.
Most existing approaches primarily rely on implicit historical pattern matching in numerical or embedding spaces, lacking explicit and interpretable reasoning mechanisms that capture latent temporal structures.
Even recent vision-enhanced time series methods~\cite{chen2024visiontsvisualmaskedautoencoders,zhong2025time,ruan2025vision} often treat visual representations as auxiliary features, rather than leveraging the ability of modern Vision-Language Models (VLMs) to perform \emph{direct reasoning on time series plots} that naturally encode trends, regime shifts, and structural transitions.
Moreover, current methods are typically designed for isolated tasks under fixed inference pipelines, with limited support for adaptive tool orchestration and coordinated execution across heterogeneous time series workloads.
These limitations can be summarized by two conceptual gaps:
\begin{itemize}[leftmargin=*, itemsep=0pt]
\item \textbf{From Representation Learning to Structural Reasoning.}
Existing methods operate in numerical or latent embedding spaces where morphological patterns remain implicit and entangled.
Recent vision-enhanced methods (e.g., TimeVLM, VisionTS) convert time series into vision primarily for \emph{representation learning}, but fail to exploit visual reasoning capabilities to extract explicit structural knowledge from plots like human analysts naturally use for temporal understanding.
\textit{Can we shift from learning representations to performing intuitive structural reasoning over time series visualizations?}

\item \textbf{From Task-Specific Models to Tool-Driven Execution.}
Current paradigms treat forecasting, classification, imputation, and anomaly detection as separate problems, typically requiring task-specific heads and individual procedures even when sharing a common backbone.
Unlike multi-agent systems that leverage tool orchestration for adaptive problem-solving, time series methods operate under the \textit{fixed model then single-task finetuning} paradigms.
It prevents pretrained models from functioning as general-purpose, task-adaptive time series solvers and necessitates multiplicative scaling costs.
\textit{Can we unify diverse time series tasks through shared structural reasoning while enabling adaptive tool selection for task-specific execution?}
\end{itemize}

General time series tasks fundamentally require an understanding of morphological semantics that are explicit in visual plots yet implicit in numerical embeddings.
This reveals that the bottleneck of time series intelligence lies in structural reasoning rather than representation learning, suggesting a paradigm shift: unify tasks through visual reasoning while enabling adaptive tool-driven execution.

Based on this insight, we propose \textbf{MAS4TS}, the first tool-driven \underline{M}ulti-\underline{A}gent \underline{S}ystem for general \underline{T}ime \underline{S}eries analysis, built upon an \textbf{Analyzer--Reasoner--Executor} paradigm.
MAS4TS decomposes time series analysis into complementary roles: the \emph{Analyzer} performs statistical analysis and data preprocessing; the \emph{Reasoner} integrates visual and numerical reasoning to extract high-level temporal priors and derive latent trajectory constraints; and the \emph{Executor} invokes appropriate tools and models to generate task-specific outputs.
Specifically, MAS4TS performs visual reasoning directly on time series plots using VLMs to obtain structured anchors, which subsequently guide trajectory reconstruction in latent space.
Through shared memory and inter-agent communication, specialized agents collaborate to solve diverse time series tasks within a unified framework.
Our key contributions are summarized as follows:
\begin{itemize}[leftmargin=*, itemsep=0pt]
\item We propose \textbf{MAS4TS}, the first tool-driven multi-agent system for general time series analysis, establishing an \textit{Analyzer--Reasoner--Executor} paradigm for task-adaptive time series modeling.
\item We introduce an intuitive reasoning approach that utilizes VLMs to extract visual anchors from plots and priors then transforms them into numerical constraints for latent trajectory reconstruction.
\item We conduct extensive experiments and demonstrate the superior performance of \model among forecasting, classification, imputation, and anomaly detection tasks.
\end{itemize}

%% file: contents/02relatedwork.tex
\section{Related Work}

\begin{figure*}[t!]
    \centering
    \includegraphics[width=0.98\textwidth]{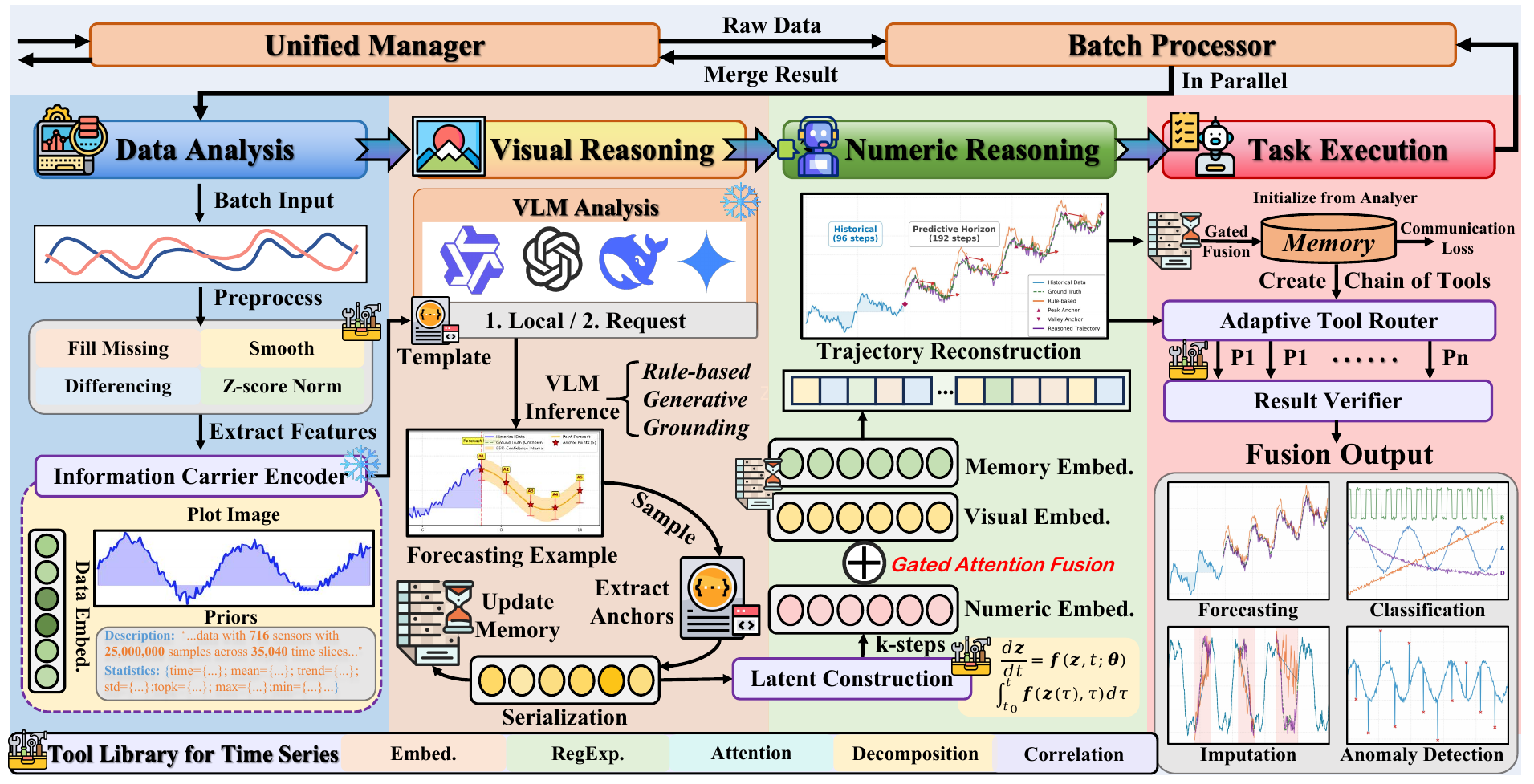}
\caption{Overview of the \model framework following an Analyzer-Reasoner-Executor paradigm. The Reasoner agent performs two sequential stages: Visual Reasoning (VLM-based anchor extraction) and Numeric Reasoning (latent trajectory reconstruction). A unified tool library supports the entire workflow, with primary integration at the Executor.}
    \label{fig:framework}
\end{figure*}

\subsection{Time Series Analysis}
Classical time series models~\cite{box2015time,cleveland1990stl} remain effective for simple or linear patterns. 
Deep learning has become the dominant paradigm in modern time series analysis, enabled by increasingly specialized architectures. 
Recurrent models~\cite{medsker2001recurrent,hochreiter1997long,cho2014learning} capture short-term temporal dependencies but struggle with gradient decay over long sequences, while convolutional models~\cite{bai2018empirical,van2016wavenet,wu2023timesnet} efficiently extract local and multi-scale patterns, offering strong computational advantages. 
Transformer-based models~\cite{zhou2024sdformer,zhou2021informer,wu2021autoformer,liu2022pyraformer,zhou2022fedformer,nie2022time} have become the mainstream due to their flexible receptive fields and modeling power, with innovations such as auto-correlation attention, frequency-domain decomposition, and patch-level representations. 
Linear architectures~\cite{zeng2023dlinear,xu2023fits,zhou2025multi} have shown surprising competitiveness on large forecasting benchmarks, renewing interest in inductive bias and architectural simplicity.

\subsection{Pretrained Large Models for Time Series}
Inspired by the success of LLMs, recent work explores foundation models for time series~\cite{liang2024foundation,jin2024position}. 
\textbf{LLM-based approaches}~\cite{jin2023time,liu2024unitime,zhou2023one,chang2025llm4ts} align time series with text embeddings, leveraging pretrained language models for cross-domain generalization. However, these methods face inherent modality gaps and computational overhead.
\textbf{Vision-enhanced approaches} convert time series to images and use pretrained vision models~\cite{chen2024visiontsvisualmaskedautoencoders,zhong2025time,shen2025multi,ruan2025vision,lyu2025occamvtsdistillingvisionmodels}, but primarily treat visual representations as auxiliary features for representation learning. 
\textit{Our work differs by leveraging VLMs to perform direct visual reasoning on time series plots, extracting explicit structural patterns that guide downstream task execution.}

\subsection{Multi-Agent Systems}
Multi-agent systems (MAS) have shown strong capabilities in complex reasoning, planning, and tool orchestration~\cite{wu2024autogen,hong2023metagpt,qian2023chatdev}. 
By assigning agents specialized roles and enabling structured communication, MAS can decompose and tackle tasks beyond the capacity of a single model. 
Tool-augmented approaches~\cite{schick2023toolformer,qin2023toolllm} further extend agent capabilities through external function invocation. 
Recent work explores agentic approaches for specific time series tasks such as forecasting~\cite{huang2025many} and anomaly detection~\cite{gu2025argos,ravuru2024agentic}, but these methods address isolated problems rather than providing a unified analytical framework. \textit{Our work fills this gap by introducing the tool-driven multi-agent framework that integrates visual reasoning for general time series analysis.}

%% file: contents/03method.tex
\section{Methodology}
\label{sec:method}
As shown in Figure~\ref{fig:framework}, \model adopts a plan-and-execute multi-agent architecture~\cite{huang2024understanding}, where a manager dynamically decomposes time series tasks into specialized stages executed by core agents. 
A processor then performs task-aware adaptive planning and enables parallel execution among agents to improve inference efficiency. 
For each parallel batch, the workflow consists of three agents: \ding{182} \emph{Analyzer}, which extracts statistical features $\mathbf{F}_{\text{stat}}$ and a task-aware plot image representation; \ding{183} \emph{Reasoner}, which leverages a VLM to generate anchors $\mathcal{A}$ (visual reasoning stage) and reconstruct the sequence in a latent space (numeric reasoning stage); and \ding{184} \emph{Executor}, which produces and verifies final predictions $\hat{\mathbf{Y}}$ through adaptive tool usage.

\subsection{Problem Formulation}
Let $\mathbf{X}=\{x_1,\dots,x_L\}\in\mathbb{R}^{L\times D}$ denote an input multivariate time series, where $L$ is the length of the observed sequence and $D$ is the number of variables (features). We consider four general tasks:
\ding{182}\emph{Forecasting} to predict future values $\mathbf{Y}=\{x_{L+1},\dots,x_{L+H}\}\in\mathbb{R}^{H\times D}$ over a forecasting horizon of length $H$;
\ding{183}\emph{Classification} to assign a categorical label $y\in\{1,\dots,C\}$ to the entire sequence, where $C$ is the number of classes;
\ding{184}\emph{Imputation} to estimate missing entries $\{\hat{x}_{i,j}:(i,j)\in\Omega\}$, with $\Omega$ denoting the index set of missing time-variable pairs; and
\ding{185}\emph{Anomaly Detection} to identify anomalous timestamps $\mathbf{s}\subset\{1,\dots,L\}$ within the observed window.
\model provides a unified agent-driven framework that shares common perception, reasoning, and memory components across these tasks.

\subsection{Multi-Agent Collaboration Framework}
\label{subsec:multi_agent}
Time series analysis requires heterogeneous reasoning that is difficult to unify in a single model, motivating our design of specialized agents that collaborate through shared memory to enable selective knowledge transfer while preserving task-specific representations.

\paragraph{Shared Memory Mechanism.}
We define a set of agents $\{\text{Agent}^{(i)}\}_{i=1}^{N}$ for batch data processing, where in our instantiation $N{=}3$ corresponds to \emph{Analyzer}, \emph{Reasoner}, and \emph{Executor}.
Each agent has (i) an embedding function $\text{Embed}^{(i)}(\cdot)$, (ii) a local memory view $\mathcal{M}^{(i)}$, and (iii) a communication module $\text{Com}(\cdot)$.
Cross-agent coordination is achieved through a shared memory matrix $\mathbf{M} \in \mathbb{R}^{L \times d_m}$, which integrates knowledge across agents via gated fusion.
Given an input time series $\mathbf{X}$ and current shared memory $\mathbf{M}$, each agent computes:
\begin{align}
\mathbf{h}^{(i)},\alpha_i &= \text{Agent}^{(i)}\big(\text{Embed}^{(i)}(\mathbf{X}), \mathcal{M}^{(i)}\big),\\
\text{Com}(\mathbf{h}^{(i)}|\mathcal{M}^{(i)}) &= \mathbf{h}^{(i)} + \text{LN}\Big(\text{Enc}_{\ell}(\mathbf{h}^{(i)}, \mathcal{M}^{(i)})\Big),\\
\mathbf{M} &\leftarrow \mathbf{M} \oplus \alpha_i \cdot \text{Com}(\mathbf{h}^{(i)}|\mathcal{M}^{(i)}),
\end{align}
where $\mathbf{h}^{(i)}$ denotes the agent-specific hidden state, $\alpha_i\in[0,1]$ is the confidence score controlling the update magnitude, and $\text{LN}(\cdot)$ denotes the layer normalization for $\ell =1,...,L_{enc}$ encoder layers conditioned on the agent view.
The gating mechanism $\oplus$ enables iterative memory refinement while reducing the impact of noisy agent outputs.

\paragraph{Information Carrier Encoder.}
Agents communicate through structured information carriers $\{\mathcal{I}, \mathcal{E}, \mathcal{P}\}$ that are first generated by the \textbf{Analyzer}, detailed as follows:
\begin{itemize}[leftmargin=*, itemsep=0pt]
\item $\mathcal{I}$: Plot images rendered directly from the raw time series via a rule-based operator $\mathcal{I} = \mathcal{R}(\{(t, x_{t,d})\}_{t=1}^{L})$;
\item $\mathcal{E}$: Normalized embeddings from tool-integrated time series preprocessing and encoding;
\item $\mathcal{P}$: Structured priors including statistical features, semantic descriptions, and shared memory from communication.
\end{itemize}
These carriers form a compact yet expressive interface for inter-agent communication and downstream reasoning.

\subsection{Latent Visual Reasoning}
\label{subsec:visual_reasoning}
The \textbf{Reasoner} aims to bridge intuitive visual pattern recognition and precise numerical inference for time series analysis.
Rather than directly fitting historical trajectories, we leverage visual reasoning to identify sparse but informative structural cues and subsequently enforce them through continuous-time latent dynamics.

\paragraph{VLM-based Anchoring.}
Many salient temporal structures in time series, including trends, regime shifts, and boundary behaviors, are often easier to identify from visual representations than from raw 1-D sequences.
Vision-Language Models (VLMs) perform high-level pattern recognition directly on plot images.
Formally, given plot images $\mathcal{I}$ and structured priors $\mathcal{P}$, we query the VLM as:
\begin{align}
\mathcal{A} &= \text{VLM}\Big(
\text{Enc}_{\text{vis}}(\mathcal{I}),\;
\text{Enc}_{\text{text}}(\text{Prompt}(\mathcal{P}))
\Big) \\ &= \{(t_k, v_k, \tau_k)\}_{k=1}^{K},
\end{align}
where the prompt (see Appendix~\ref{appx:prompt}) specifies the required output schema.
Each tuple consists of a critical time index $t_k \in \{1,\dots,L\}$, an estimated value $v_k \in \mathbb{R}$, and an indicator $\tau_k$ (default: $\{-1,0,1\}$) denoting local temporal behavior (e.g., falling/stable/rising).
These sparse anchors encode high-level structural knowledge extracted through visual reasoning, serving as semantic constraints that guide the latent reconstruction.

\paragraph{Latent Construction.}
While visual anchors are sparse and discrete, downstream time series tasks require dense and temporally coherent representations.
To bridge this gap, we introduce a \emph{latent construction} module that reconstructs a continuous latent trajectory by treating anchors as conditioning signals that guide latent-space evolution.
Formally, we define a continuous-time latent state $\mathbf{z}(t)$ and parameterize its dynamics by:
\begin{equation}
\frac{d\mathbf{z}(t)}{dt}
= f_{\phi}\big(\mathbf{z}(t), t \mid \mathcal{A}, \mathcal{P}\big),
\quad \mathbf{z}(t_0) = \mathbf{z}_0,
\end{equation}
where $f_{\phi}$ is a learnable dynamics function conditioned on the anchor set $\mathcal{A}$ and priors $\mathcal{P}$, and the initial state $\mathbf{z}_0$ is obtained from the embedding $\mathcal{E}$. We employ a finite-step latent reconstruction operator $\mathcal{S}(\cdot)$ to obtain latent states
$\{\mathbf{z}_t\}_{t=1}^{L+H}$, covering the observed window and the horizon:
\begin{equation}
\mathbf{u}^{(k)} = \mathcal{S}^{(k)}\!\big(\cdots\mathcal{S}^{(1)}(\mathbf{u}^{(0)}, \mathcal{A}, \mathcal{P})\big), \quad \mathbf{z}_t = [\mathbf{u}^{(k)}]_t,
\end{equation}
where $\mathbf{u}^{(i)} \in \mathbb{R}^{(L+H)\times D}$ denotes the \emph{sequence-level} latent representation after the $i$-th solver step, with $\mathbf{u}^{(0)}=\text{Proj}(\mathcal{E})$ initialized from the data embedding and fused by communication, and $\mathbf{z}_t$ being the $t$-th row of $\mathbf{u}^{(k)}$.

In practice, we instantiate $\mathcal{S}(\cdot)$ using a finite-step ($k \leq 20$) explicit latent ODE solver, as anchor conditioning substantially restricts the feasible solution space and reduces the effective curvature of the latent trajectory.
Intuitively, and consistent with prior work on continuous latent modeling~\cite{dormand1980family,rubanova2019latent},
anchor-conditioned latent dynamics act as a form of \emph{soft boundary regularization}, biasing the reconstructed trajectory toward semantically meaningful temporal structures while preserving flexibility in local variations.

\paragraph{Latent Fusion with Data Embedding.}
The reconstructed latent trajectory provides high-level structural guidance but does not replace the original data representation.
Instead, it is fused with the data embedding to form a unified latent representation using attention fusion:
\begin{equation}
\mathbf{Z} = \text{Attn}(\mathcal{E} \;\oplus\; \text{Proj}\!\left( \mathbf{u}^{(k)} \right)),
\end{equation}
where $\oplus$ denotes a gated fusion operator.
The resulting latent representation $\mathbf{Z}$ is dense, differentiable, and semantically grounded, serving as a bridge between training-free visual reasoning and data-driven numerical inference within a unified latent framework.

\subsection{Tool-Integrated Execution and Optimization}
\label{sec:execution_optimization}
The \textbf{Executor} converts latent representations into task-specific outputs via tool-integrated execution and result verification, while optimizing only coordination-critical components and keeping the reasoning modules training-free.
Instead of relying on a single monolithic predictor, \model dynamically composes a chain-of-tools that applies specialized time-series operators to refine intermediate results.

\paragraph{Tool Library for Time Series.}
At execution time, an adaptive router selects a tool chain from a predefined library $\mathcal{T}=\{\mathcal{T}_1,\dots,\mathcal{T}_M\}$, where each tool implements a specialized operation (e.g., trend extrapolation, decomposition, smoothing, statistical prediction, constraint projection, and residual correction).
Tool routing adapts to the task type $\kappa$, structured priors $\mathcal{P}$, and the shared memory $\mathbf{M}$ maintained in earlier stages. 
Formally, the executor produces a routing policy $\pi_{\Theta}$ over tools and constructs a tool sequence $\mathbf{m}=(m_1,\dots,m_J)$ accordingly. The final output is obtained by sequential composition:
\begin{equation}
\pi_{\Theta}(m \mid \kappa, \mathcal{P}, \mathbf{M})
= \mathrm{Softmax}\!\left(g_{\psi}(\kappa, \mathcal{P}, \mathbf{M})\right),
\end{equation}
\begin{equation}
\hat{\mathbf{Y}}
= \mathcal{T}_{m_J} \circ \cdots \circ \mathcal{T}_{m_1}(\mathbf{Z}),
\end{equation}
where $g_{\psi}$ is a gated controller network, and the tool sequence with length $J$ is dynamically determined by policy $\pi_{\Theta}$ and connected by $\circ$ operation.
This design enhances raw predictions through plug-and-play expert operators without retraining a monolithic backbone for a specific task.

\paragraph{Result Verifier.}
To ensure numerical validity and task consistency, \model employs a lightweight verifier that performs:
(i) format and shape correction,
(ii) constraint validation, and
(iii) task-specific normalization.
If the candidate output fails verification, the executor either recomputes or falls back to a conservative basic tool for a valid prediction.
Given a candidate output $\hat{\mathbf{Y}}$, the verifier produces the final prediction:
\begin{equation}
\mathbf{Y}^{*} = \text{Verifier}(\hat{\mathbf{Y}} \mid \mathcal{A}, \mathcal{P}),
\end{equation}
where visual anchors $\mathcal{A}$ and structured priors $\mathcal{P}$ serve as external consistency checks.
For continuous outputs, violations of anchor-implied bounds are softly corrected via constraint-aware projection; for discrete or categorical tasks, the verifier enforces schema consistency and removes logically invalid predictions.
This decouples high-level reasoning from strict constraint satisfaction, improving robustness without sacrificing flexibility.

\paragraph{Optimization Strategy.}
\model supports end-to-end differentiable execution while selectively training only coordination-critical components.
Static preprocessing, VLM prompting/inference, and selected tools in $\mathcal{T}$ are kept \emph{training-free}, serving as fixed reasoning and execution primitives.
Optimization focuses on learnable components that govern information flow and coordination, including:
(i) embedding/communication functions,
(ii) shared memory representations,
and (iii) gating and routing parameters $\Theta$.
We optimize a task-dependent objective:
\begin{equation}
\mathcal{L}
= \mathcal{L}_{\text{task}}(\mathbf{Y}^{*}, \mathbf{Y})
+ \lambda \cdot \mathcal{L}_{\text{com}}(\mathbf{M}),
\end{equation}
where $\mathcal{L}_{\text{task}}$ corresponds to the downstream objective (e.g., $\mathcal{L}_{\mathrm{MSE}}$ for forecasting),
and $\mathcal{L}_{\text{com}}$ regularizes memory updates to encourage stable inter-agent communication.

\model integrates multi-agent collaboration, intuitive visual reasoning, and tool-augmented execution into a unified framework.
By explicitly separating structural reasoning, continuous latent reconstruction, and tool-integrated execution, \model achieves strong adaptability across diverse time series tasks while preserving numerical consistency and training efficiency.

%% file: contents/04experiments.tex
\input{tables/forecasting}
\input{tables/classification}

\section{Experiments}
\label{sec:experiments}
We conduct extensive experiments to evaluate \model on four fundamental time series tasks: forecasting, classification, imputation, and anomaly detection.
We benchmark on 21 widely-used datasets and compare against more than 20 competitive baselines, including task-specific time series models and recent LM-based approaches.
For fair comparison, we follow standard protocols for each task (data splits, input/output lengths, and evaluation metrics) and use the same splits and evaluation settings across methods.
Experimental details are provided in Appendix~\ref{appx:experiment_details}.

\subsection{Forecasting}
\label{sec:exp_forecasting}
\noindent\textbf{Setups.}
For long-term forecasting, we evaluate on 7 widely-adopted benchmarks: ETT (4 subsets: ETTh1, ETTh2, ETTm1, ETTm2), Weather, Solar-Energy, and Exchange-Rate.
Following established protocols~\cite{wu2023timesnet,liu2023itransformer}, we use a fixed look-back window of 96 and evaluate prediction horizons \(H \in \{96, 192, 336, 720\}\).

\noindent\textbf{Results.} As shown in Table~\ref{tab:long_term_forecasting_results}, \model achieves competitive performance across all benchmarks. On the Weather dataset, \model obtains an average MSE of 0.232, outperforming MAFS (0.252) and TimeVLM (0.262) by 7.9\% and 11.5\%, respectively.
Temporal dependencies exhibit non-stationary patterns with regime shifts on the Exchange dataset, while \model achieves 0.343 average MSE, demonstrating that visual anchoring effectively captures abrupt trend changes that challenge purely numerical approaches. 
The ETT benchmarks further validate the robustness of our framework: \model consistently ranks among the top performers across all four subsets, with particularly strong results on ETTh2 (0.379 MSE), where complex multi-scale patterns require both fine-grained numerical reasoning and high-level structural understanding.

\subsection{Classification}
\noindent\textbf{Setups.} We evaluate on 8 multivariate datasets from the UEA archive~\cite{bagnall2018uea}: EthanolConcentration (chemical), FaceDetection (video), Handwriting (motion capture), Heartbeat (medical), JapaneseVowels (audio), SelfRegulationSCP1/SCP2 (EEG), and UWaveGestureLibrary (gesture recognition). These datasets span diverse domains with varying sequence lengths (from 29 to 1,751), dimensionality (from 3 to 144), and class counts.

\noindent\textbf{Results.} Table~\ref{tab:classification} presents the classification results. \model achieves the highest average accuracy of 68.25\%, outperforming iTransformer (65.16\%) and DLinear (64.85\%) by 4.7\% and 5.2\%, respectively. On FaceDetection, which requires capturing subtle temporal dynamics in video sequences, \model achieves 69.38\% accuracy compared to 65.3\% for TiDE, demonstrating the discriminative power of VLM-extracted visual features. \model maintains robust performance across datasets with heterogeneous characteristics, whereas specialized methods like Informer show higher variance.

\subsection{Imputation}
\noindent\textbf{Setups.} We evaluate imputation on the ETT family, Electricity, Weather, and Exchanges datasets under random missing scenarios with mask ratios $r \in \{12.5\%, 25\%, 37.5\%, 50\%\}$. Missing values are randomly sampled across all time steps and variables, following the protocol in~\cite{wu2023timesnet}.

\noindent\textbf{Results.} As shown in Table~\ref{tab:imputation}, \model achieves state-of-the-art imputation performance across all mask ratios. 
At 50\% missing ratio on ETTm2, \model obtains an MSE of 0.030, reducing the error by 34.8\% compared to the second LightTS (0.046). 
The improvement margin increases with higher mask ratios, demonstrating strong robustness under severe data corruption.
On the Exchange dataset, \model achieves an average of 0.005 MSE, outperforming all the baselines.
This validates that our anchor-constrained completion is particularly effective when missing segments are extensive: visual anchors provide reliable structural waypoints, while the latent construction treats these anchors as boundary constraints to produce smooth and temporally coherent trajectories. 
Although gains on low missing ratios are more modest, the multi-agent coordination mechanism ensures consistent improvements, and the foundation model's generalization capability offers potential for further enhancement on domain-specific missing patterns.

\subsection{Anomaly Detection}
\noindent\textbf{Setups.} We evaluate on five server and infrastructure monitoring benchmarks: SMD~\cite{su2019robust} (server machine dataset with 38 dimensions), MSL and SMAP~\cite{hundman2018detecting} (spacecraft telemetry from NASA), SWaT~\cite{mathur2016swat} (water treatment plant with 51 sensors), and PSM~\cite{abdulaal2021practical} (server metrics from eBay).
These datasets contain both point anomalies and contextual anomalies with varying contamination ratios.

\input{tables/imputation}
\input{tables/anomaly}
\noindent\textbf{Results.} Table~\ref{tab:anomaly} presents the anomaly detection results. \model achieves the highest average F1-score of 84.78\%, improving over ETSformer (82.87\%) and iTransformer (76.98\%) by 1.5\% and 9.3\%, respectively.
On SWaT, \model achieves 93.14\% F1-score with balanced precision (91.61\%) and recall (94.74\%), indicating effective discrimination between normal fluctuations and genuine anomalies.
The \textbf{Reasoner} with visual reasoning proves particularly valuable for anomaly detection: by directly ``seeing'' anomalous patterns from time series visualizations, it can identify salient deviations before the model fully converges, pre-locating suspicious regions that guide subsequent numerical analysis. 
Visual anchors encode expected temporal structures learned from normal patterns, enabling sensitive detection when observations deviate from these structural priors. 
In addition, the \textbf{Executor}'s verification step filters false positives via anchor-consistency checks, improving precision without sacrificing recall.

\begin{figure}[!ht]
\centering
\includegraphics[width=0.95\linewidth]{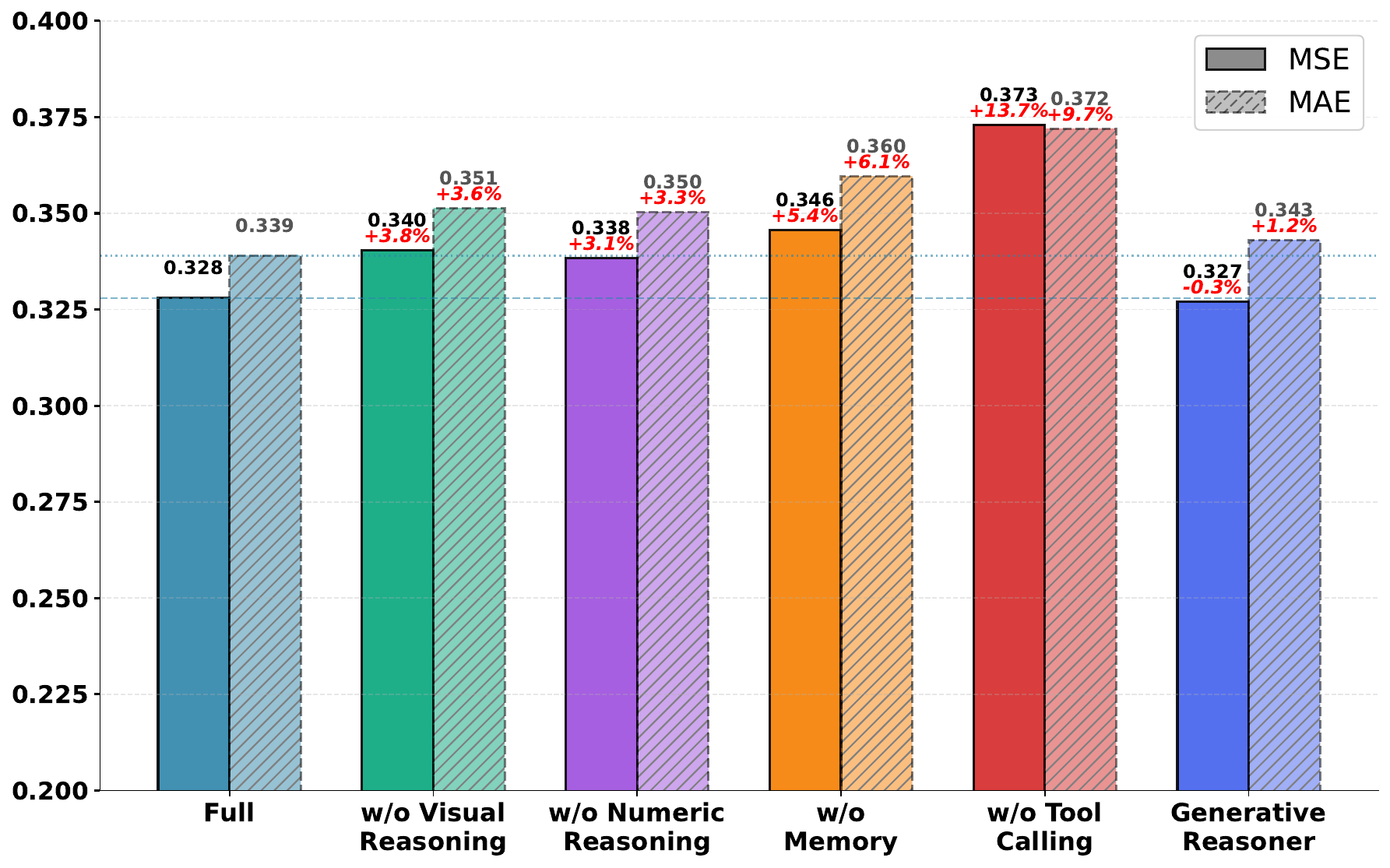}
\caption{The ablation results on the Weather dataset with prediction horizon $H=720$. Each component contributes to the performance, while generative reasoner does not improve further.}
\label{fig:ablation}
\end{figure}

\subsection{Model Analysis}
\paragraph{Ablation Studies.}
\label{subsec:ablation}

We conduct systematic ablation studies to quantify the contribution of key components in \model.
All variants are evaluated on the Weather dataset with prediction horizon \(H=720\).
Figure~\ref{fig:ablation} reports results for the full \model and the following variants:
(a) without Visual Reasoning;
(b) without Numeric Reasoning;
(c) without Shared Memory;
(d) without Tools and use a simple predictor in the \textbf{Executor}; and
(e) \textit{Generative Reasoner}, which replaces the default rule-based rendering pipeline with a generative model (\textit{Nano-Banana}) to generate the visualization, followed by the same anchor extraction procedure.

The ablation results reveal several insights.
Removing Tool Calling causes the largest performance degradation, increasing MSE by 13.7\% (0.328 \(\rightarrow\) 0.373) and MAE by 9.7\% (0.339 \(\rightarrow\) 0.372), highlighting that adaptive tool invocation and integration are crucial for handling forecasting scenarios. 
Disabling Visual Reasoning, Numeric Reasoning, or Shared Memory leads to a certain degree of MSE increases (3.1\%--5.4\%), indicating that these components provide complementary signals and collectively improve robustness.
The \textit{Generative Reasoner} variant does not yield a clear improvement over the default instruct-style anchor extraction; nevertheless, this result does not preclude the potential of directly reasoning with generated visualizations under more suitable settings.
Overall, these findings suggest that \model benefits from the coordinated use of multiple reasoning modalities, where tool adaptation plays a central role in effectively leveraging the available components.

\begin{figure}[!ht]
    \centering
    \includegraphics[width=\linewidth]{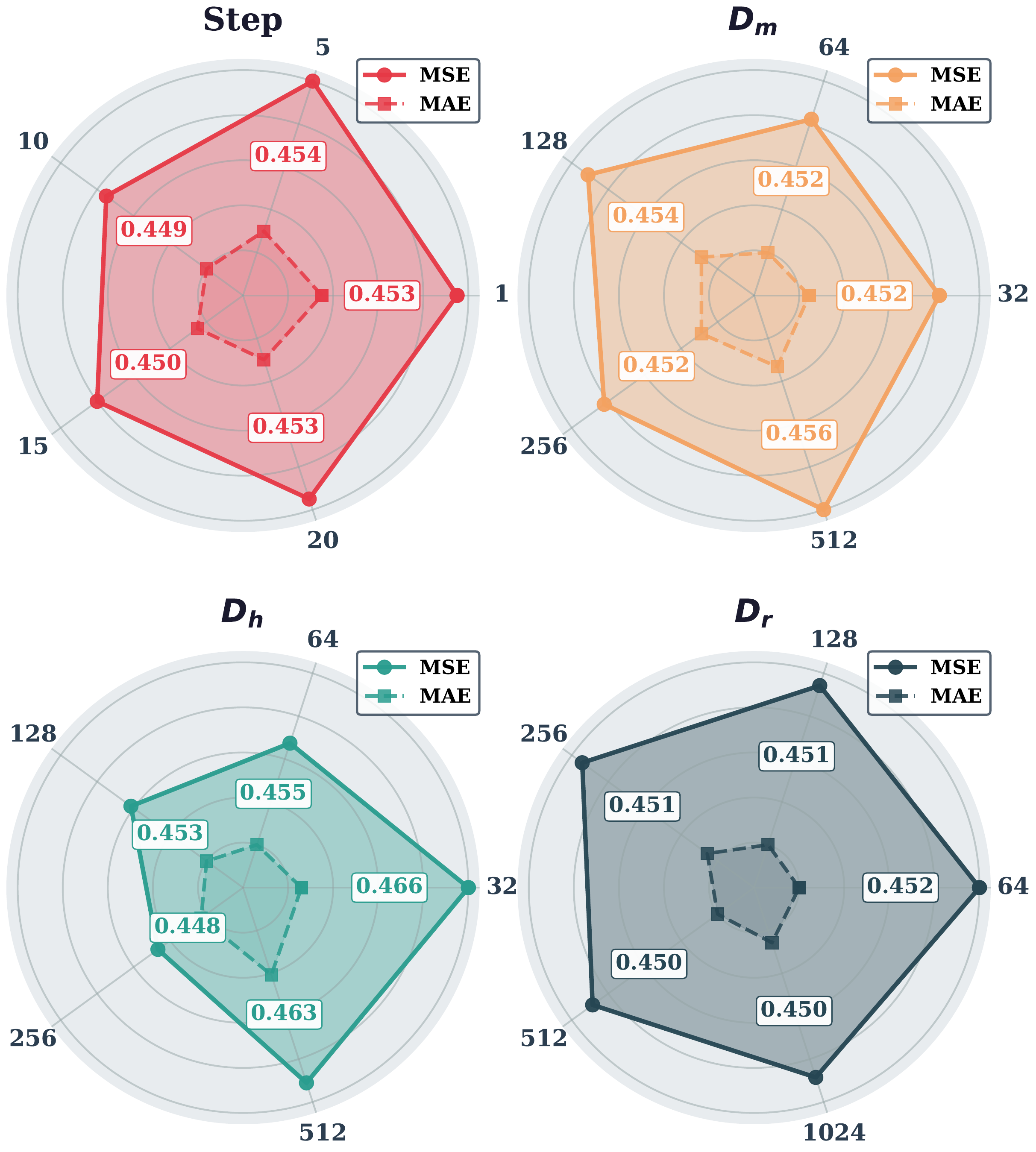}
    \caption{Hyperparameter sensitivity analysis on ETTm1 ($H=720$). Each radar chart shows MSE (solid) and MAE (dashed) for: a) reasoning steps; b) memory dimension $D_m$; c) hidden dimension $D_h$; and d) feedforward dimension $D_r$.}
    \label{fig:hyperparameter}
\end{figure}

\paragraph{Hyperparameter Studies.}
We analyze the impact of key hyperparameters on ETTm1 with prediction horizon $H=720$, as shown in Figure~\ref{fig:hyperparameter}. The radar charts visualize MSE and MAE variations across different parameter settings.

\ding{182}\textit{Reasoning Steps}: Performance improves as steps increase from 1 to 20, achieving an optimal MSE of 0.449 at 10 steps, then slightly degrades with further increases. This suggests that 10 integration steps provide sufficient numerical precision for the latent ODE-solver while avoiding computational overhead. \ding{183}\textit{Dimension of Communication Memory $D_m$}: The model exhibits remarkable robustness to memory dimension variations, with MSE ranging narrowly from 0.452 to 0.456 across all settings (32-512). We adopt $D_m=64$ as the default for computational efficiency. \ding{184}\textit{Dimension of Hidden State $D_h$}: Performance consistently improves as $D_h$ increases from 32 (MSE=0.466) to 256 (MSE=0.448), but degrades at 512 due to overfitting. The optimal $D_h=256$ balances model capacity and generalization. 
\ding{185}\textit{Feedforward Dimension $D_r$}: Similar to $D_m$, the model shows low sensitivity to $D_r$, with MSE stabilizing around 0.450 for $D_r \geq 128$. We set $D_r=512$ as the default, which achieves marginally better performance without high computational cost.

\input{appendix/efficiency}

%% file: tables/forecasting.tex
\begin{table*}[!ht]
  \caption{Summarized results for the long-term forecasting task. We report the averaged metrics (\emph{Avg}) across four prediction lengths $\{96, 192, 336, 720\}$ for each dataset. Full results are available in Table~\ref{tab:full_long_term_forecasting_results}. \boldres{Red bold}: the best results; \secondres{blue}: the second best.}
  \label{tab:long_term_forecasting_results}
  \centering
  \resizebox{1.0\linewidth}{!}{
  \begin{threeparttable}
  \begin{small}
  \renewcommand{\multirowsetup}{\centering}
  \setlength{\tabcolsep}{1.2pt}
  \begin{tabular}{c|cc|cc|cc|cc|cc|cc|cc|cc|cc|cc|cc|cc|cc|cc}
    \toprule
    \multicolumn{1}{c}{\multirow{2}{*}{Models}} & \multicolumn{2}{c}{\rotatebox{0}{\scalebox{0.8}{\textbf{MAS4TS}}}} &
    \multicolumn{2}{c}{\rotatebox{0}{\scalebox{0.8}{MAFS}}} &
    \multicolumn{2}{c}{\rotatebox{0}{\scalebox{0.8}{TimeVLM}}} &
    \multicolumn{2}{c}{\rotatebox{0}{\scalebox{0.8}{UniTime}}} &
    \multicolumn{2}{c}{\rotatebox{0}{\scalebox{0.8}{iTransformer}}} &
    \multicolumn{2}{c}{\rotatebox{0}{\scalebox{0.8}{PatchTST}}} &
    \multicolumn{2}{c}{\rotatebox{0}{\scalebox{0.8}{Crossformer}}} &
    \multicolumn{2}{c}{\rotatebox{0}{\scalebox{0.8}{TiDE}}} &
    \multicolumn{2}{c}{\rotatebox{0}{\scalebox{0.8}{TimesNet}}} & 
    \multicolumn{2}{c}{\rotatebox{0}{\scalebox{0.8}{DLinear}}} &
    \multicolumn{2}{c}{\rotatebox{0}{\scalebox{0.8}{SCINet}}}&
    \multicolumn{2}{c}{\rotatebox{0}{\scalebox{0.8}{FEDformer}}} & 
    \multicolumn{2}{c}{\rotatebox{0}{\scalebox{0.8}{Stationary}}} & 
    \multicolumn{2}{c}{\rotatebox{0}{\scalebox{0.8}{Autoformer}}}  \\
    \multicolumn{1}{c}{} & \multicolumn{2}{c}{\scalebox{0.8}{(\textbf{Ours})}} &
    \multicolumn{2}{c}{\scalebox{0.8}{\citeyear{huang2025many}}} &
    \multicolumn{2}{c}{\scalebox{0.8}{\citeyear{zhong2025time}}} &
    \multicolumn{2}{c}{\scalebox{0.8}{\citeyear{liu2024unitime}}} &
    \multicolumn{2}{c}{\scalebox{0.8}{\citeyear{liu2023itransformer}}} &
    \multicolumn{2}{c}{\scalebox{0.8}{\citeyear{nie2023patchtst}}}&
    \multicolumn{2}{c}{\scalebox{0.8}{\citeyear{zhang2023crossformer}}}&
    \multicolumn{2}{c}{\scalebox{0.8}{\citeyear{das2023long}}}&
    \multicolumn{2}{c}{\scalebox{0.8}{\citeyear{wu2022timesnet}}}&
    \multicolumn{2}{c}{\scalebox{0.8}{\citeyear{zeng2023dlinear}}}&
    \multicolumn{2}{c}{\scalebox{0.8}{\citeyear{liu2022scinet}}}&
    \multicolumn{2}{c}{\scalebox{0.8}{\citeyear{zhou2022fedformer}}}&
    \multicolumn{2}{c}{\scalebox{0.8}{\citeyear{liu2022non}}}&
    \multicolumn{2}{c}{\scalebox{0.8}{\citeyear{wu2021autoformer}}}
    \\
    
    \cmidrule(lr){2-3} \cmidrule(lr){4-5} \cmidrule(lr){6-7} \cmidrule(lr){8-9} \cmidrule(lr){10-11} \cmidrule(lr){12-13} \cmidrule(lr){14-15} \cmidrule(lr){16-17} \cmidrule(lr){18-19} \cmidrule(lr){20-21} \cmidrule(lr){22-23} \cmidrule(lr){24-25} \cmidrule(lr){26-27} \cmidrule(lr){28-29}
    \multicolumn{1}{c}{Metric} & \scalebox{0.78}{MSE} & \scalebox{0.78}{MAE} & \scalebox{0.78}{MSE} & \scalebox{0.78}{MAE} & \scalebox{0.78}{MSE} & \scalebox{0.78}{MAE} & \scalebox{0.78}{MSE} & \scalebox{0.78}{MAE} & \scalebox{0.78}{MSE} & \scalebox{0.78}{MAE} & \scalebox{0.78}{MSE} & \scalebox{0.78}{MAE} & \scalebox{0.78}{MSE} & \scalebox{0.78}{MAE} & \scalebox{0.78}{MSE} & \scalebox{0.78}{MAE} & \scalebox{0.78}{MSE} & \scalebox{0.78}{MAE} & \scalebox{0.78}{MSE} & \scalebox{0.78}{MAE} & \scalebox{0.78}{MSE} & \scalebox{0.78}{MAE} & \scalebox{0.78}{MSE} & \scalebox{0.78}{MAE} & \scalebox{0.78}{MSE} & \scalebox{0.78}{MAE} & \scalebox{0.78}{MSE} & \scalebox{0.78}{MAE}\\
    \toprule

    \scalebox{0.85}{Weather} &\boldres{\scalebox{0.85}{0.232}}&\boldres{\scalebox{0.85}{0.270}}& \secondres{\scalebox{0.78}{0.252}} & \secondres{\scalebox{0.78}{0.275}} & \scalebox{0.78}{0.262} & \scalebox{0.78}{0.281} & \scalebox{0.78}{0.253} & \scalebox{0.78}{0.276} & {\scalebox{0.78}{0.258}} & {\scalebox{0.78}{0.278}}&{\scalebox{0.85}{0.265}}&\scalebox{0.85}{0.285}&\scalebox{0.85}{0.264}&\scalebox{0.85}{0.320}& \scalebox{0.78}{0.271} & \scalebox{0.78}{0.320} &{\scalebox{0.78}{0.259}} &{\scalebox{0.78}{0.287}}&\scalebox{0.85}{0.265}&\scalebox{0.85}{0.315}&\scalebox{0.85}{0.292} &\scalebox{0.85}{0.363} &\scalebox{0.85}{0.309} &\scalebox{0.85}{0.360} &\scalebox{0.85}{0.288} &\scalebox{0.85}{0.314} &\scalebox{0.85}{0.338} &\scalebox{0.85}{0.382}  \\
    \midrule
    
    \scalebox{0.85}{Solar-Energy}&\boldres{\scalebox{0.8}{0.212}} &\scalebox{0.8}{0.276} & \scalebox{0.78}{0.234} & \secondres{\scalebox{0.78}{0.265}} & \scalebox{0.78}{0.259} & \scalebox{0.78}{0.293} & \scalebox{0.78}{0.270} & \scalebox{0.78}{0.315} &\secondres{\scalebox{0.78}{0.233}} &\boldres{\scalebox{0.78}{0.262}}&\scalebox{0.85}{0.287}&\scalebox{0.85}{0.333}&\scalebox{0.85}{0.406}&\scalebox{0.85}{0.442}&\scalebox{0.78}{0.347} &\scalebox{0.78}{0.417}&\scalebox{0.85}{0.403}&\scalebox{0.85}{0.374}&\scalebox{0.85}{0.330}&\scalebox{0.85}{0.401}&\scalebox{0.85}{0.282} &\scalebox{0.85}{0.375}&\scalebox{0.85}{0.328} &\scalebox{0.85}{0.383} &\scalebox{0.85}{0.350} &\scalebox{0.85}{0.390} &\scalebox{0.85}{0.586} &\scalebox{0.85}{0.557} \\
    \midrule

    \scalebox{0.85}{Exchange} & \boldres{\scalebox{0.8}{0.343}} & \boldres{\scalebox{0.8}{0.397}} & \scalebox{0.78}{0.396} & \scalebox{0.78}{0.417} & \secondres{\scalebox{0.78}{0.355}} & \secondres{\scalebox{0.78}{0.399}} & \scalebox{0.78}{0.465} & \scalebox{0.78}{0.462} &{\scalebox{0.78}{0.360}} & {\scalebox{0.78}{0.403}} & \scalebox{0.85}{0.367} & {\scalebox{0.85}{0.404}} & \scalebox{0.85}{0.940} & \scalebox{0.85}{0.707} & \scalebox{0.78}{0.370} & \scalebox{0.78}{0.413} & \scalebox{0.78}{0.416} & \scalebox{0.78}{0.443} & {\scalebox{0.85}{0.354}} & \scalebox{0.85}{0.414} & \scalebox{0.85}{0.750} & \scalebox{0.85}{0.626} & \scalebox{0.85}{0.519} & \scalebox{0.85}{0.429} & \scalebox{0.85}{0.461} & \scalebox{0.85}{0.454} & \scalebox{0.85}{0.613} & \scalebox{0.85}{0.539} \\
    \midrule

    \scalebox{0.85}{ETTh1} &\boldres{\scalebox{0.85}{0.440}}&\boldres{\scalebox{0.85}{0.438}}& \scalebox{0.78}{0.450} & \secondres{\scalebox{0.78}{0.440}} & \scalebox{0.78}{0.457} & \scalebox{0.78}{0.443} & \secondres{\scalebox{0.78}{0.442}} & \scalebox{0.78}{0.448} & {\scalebox{0.78}{0.454}} & {\scalebox{0.78}{0.447}}&\scalebox{0.85}{0.516}&{\scalebox{0.85}{0.484}}&\scalebox{0.85}{0.529}&\scalebox{0.85}{0.522}& \scalebox{0.78}{0.541} & \scalebox{0.78}{0.507} &\scalebox{0.78}{0.458} &{\scalebox{0.78}{0.450}}&{\scalebox{0.85}{0.461}}&\scalebox{0.85}{0.457}&\scalebox{0.85}{0.747} &\scalebox{0.85}{0.647}&\scalebox{0.85}{0.498} &\scalebox{0.85}{0.484} &\scalebox{0.85}{0.570} &\scalebox{0.85}{0.537} &\scalebox{0.85}{0.496} &\scalebox{0.85}{0.487} \\
    \midrule

    \scalebox{0.85}{ETTh2} &\boldres{\scalebox{0.85}{0.379}}&\boldres{\scalebox{0.85}{0.402}}& \scalebox{0.78}{0.380} & \secondres{\scalebox{0.78}{0.403}} & \scalebox{0.78}{0.384} & \scalebox{0.78}{0.405} & \secondres{\scalebox{0.78}{0.379}} & \scalebox{0.78}{0.403} & {\scalebox{0.78}{0.383}} & {\scalebox{0.78}{0.407}}&\scalebox{0.85}{0.391}&\scalebox{0.85}{0.411}&\scalebox{0.85}{0.942}&\scalebox{0.85}{0.684}& \scalebox{0.78}{0.611} & \scalebox{0.78}{0.550}  &{\scalebox{0.78}{0.414}} &{\scalebox{0.78}{0.427}}&\scalebox{0.85}{0.563}&\scalebox{0.85}{0.519} &\scalebox{0.85}{0.954} &\scalebox{0.85}{0.723}&\scalebox{0.85}{0.437} &\scalebox{0.85}{0.449} &\scalebox{0.85}{0.526} &\scalebox{0.85}{0.516} &\scalebox{0.85}{0.450} &\scalebox{0.85}{0.459} \\
    \midrule

    \scalebox{0.85}{ETTm1} &\boldres{\scalebox{0.85}{0.367}}&\boldres{\scalebox{0.85}{0.381}}& \scalebox{0.78}{0.397} & \scalebox{0.78}{0.402} & \scalebox{0.78}{0.388} & \secondres{\scalebox{0.78}{0.398}} & \secondres{\scalebox{0.78}{0.385}} & \scalebox{0.78}{0.399} & \scalebox{0.78}{0.407} & \scalebox{0.78}{0.410}&{\scalebox{0.85}{0.406}}&\scalebox{0.85}{0.407}&\scalebox{0.85}{0.513}&\scalebox{0.85}{0.495}& \scalebox{0.78}{0.419} & \scalebox{0.78}{0.419} &{\scalebox{0.78}{0.400}} &{\scalebox{0.78}{0.406}}&\scalebox{0.85}{0.404}&\scalebox{0.85}{0.408} &\scalebox{0.85}{0.485} &\scalebox{0.85}{0.481} &\scalebox{0.85}{0.448} &\scalebox{0.85}{0.452} &\scalebox{0.85}{0.481} &\scalebox{0.85}{0.456} &\scalebox{0.85}{0.588} &\scalebox{0.85}{0.517} \\
    \midrule

    \scalebox{0.85}{ETTm2} &\boldres{\scalebox{0.85}{0.278}}&\boldres{\scalebox{0.85}{0.326}}& \scalebox{0.78}{0.288} & \scalebox{0.78}{0.330} & \secondres{\scalebox{0.78}{0.279}} & \secondres{\scalebox{0.78}{0.329}} & \scalebox{0.78}{0.293} & \scalebox{0.78}{0.334} & {\scalebox{0.78}{0.288}} & {\scalebox{0.78}{0.332}}&\scalebox{0.85}{0.290}&\scalebox{0.85}{0.334}&\scalebox{0.85}{0.757}&\scalebox{0.85}{0.610}& \scalebox{0.78}{0.358} & \scalebox{0.78}{0.404} &{\scalebox{0.78}{0.291}} &{\scalebox{0.78}{0.333}}&\scalebox{0.85}{0.354}&\scalebox{0.85}{0.402} &\scalebox{0.85}{0.954} &\scalebox{0.85}{0.723}&\scalebox{0.85}{0.305} &\scalebox{0.85}{0.349} &\scalebox{0.85}{0.306} &\scalebox{0.85}{0.347} &\scalebox{0.85}{0.327} &\scalebox{0.85}{0.371} \\
\bottomrule
  \end{tabular}
    \end{small}
  \end{threeparttable}
  }
\end{table*}

%% file: tables/classification.tex
\begin{table*}[!ht]
\caption{Average classification accuracy (\%) on 8 UEA multivariate time series datasets. \boldres{Red bold}: the best result; \secondres{blue}: the second best. The full results among all datasets are provided in Table~\ref{tab:full_classification_results}.}
\centering
\label{tab:classification}
  \resizebox{\linewidth}{!}{
  \setlength{\tabcolsep}{3pt}
  \begin{tabular}{c|cccccccccccccccccc}
    \toprule
    \multicolumn{1}{c|}{\multirow{2}{*}{Models}} & {DTW} & {XGBoost} & {LSTM} & {LSTNet} & {LSSL} & {TCN} & {Trans.} & {Re.} & {In.} & {Pyra.} & {Auto.} & {FED.} & {ETS.} & {iTrans.} & {DLinear} & {LightTS} & {TiDE} & {\textbf{MAS4TS}} \\
	 & {\citeyearpar{Berndt1994UsingDT}} & {\citeyearpar{Chen2016XGBoostAS}} & {\citeyearpar{Hochreiter1997LongSM}} & 
	{\citeyearpar{2018Modeling}} & 
	{\citeyearpar{gu2022efficiently}} & 
	{\citeyearpar{Franceschi2019UnsupervisedSR}} & {\citeyearpar{vaswani2017attention}} & 
	{\citeyearpar{kitaev2020reformer}} & {\citeyearpar{zhou2021informer}} & {\citeyearpar{liu2022pyraformer}} &
	{\citeyearpar{wu2021autoformer}} & 
	{\citeyearpar{zhou2022fedformer}} & {\citeyearpar{woo2022etsformer}} & {\citeyearpar{liu2023itransformer}} & 
	{\citeyearpar{zeng2023dlinear}} & {\citeyearpar{campos2023lightts}} & {\citeyearpar{das2023long}}& {\textbf{(Ours)}} \\
    \midrule
    Avg. Acc. & {62.78} & {61.48} & {51.72} & {66.40} & {65.30} & {67.26} & \secondres{{67.31}} & {66.93} & {59.93} & {65.58} & {66.00} & {65.32} & {65.83} & {65.16} & {64.85} & {64.47} & {64.16} & \boldres{{68.25}} \\
    \bottomrule
  \end{tabular}
  }
\end{table*}

%% file: tables/imputation.tex
\begin{table*}[!ht]
\centering
\caption{Results for the imputation task. We randomly mask 12.5\%, 25\%, 37.5\% and 50\% time points and report the average performance. Full results are available in Table~\ref{tab:full_imputation_results}. The best and second-best results are \boldres{red} and \secondres{blue}, respectively.}
\label{tab:imputation}
\resizebox{\textwidth}{!}{
\begin{tabular}{c|cc|cc|cc|cc|cc|cc|cc|cc|cc|cc|cc}
\hline
\multirow{2}{*}{Models} & \multicolumn{2}{c|}{MAS4TS} & \multicolumn{2}{c|}{ETSformer} & \multicolumn{2}{c|}{LightTS} & \multicolumn{2}{c|}{DLinear} & \multicolumn{2}{c|}{FEDformer} & \multicolumn{2}{c|}{Pyraformer} & \multicolumn{2}{c|}{Informer} & \multicolumn{2}{c|}{Reformer} & \multicolumn{2}{c|}{LSTM} & \multicolumn{2}{c|}{TCN} & \multicolumn{2}{c}{LSSL} \\
& \multicolumn{2}{c|}{(Ours)} & \multicolumn{2}{c|}{(2022)} & \multicolumn{2}{c|}{(2022)} & \multicolumn{2}{c|}{(2023)} & \multicolumn{2}{c|}{(2022)} & \multicolumn{2}{c|}{(2021a)} & \multicolumn{2}{c|}{(2021)} & \multicolumn{2}{c|}{(2020)} & \multicolumn{2}{c|}{(1997)} & \multicolumn{2}{c|}{(2019)} & \multicolumn{2}{c}{(2022)} \\
\hline
Metric & MSE & MAE & MSE & MAE & MSE & MAE & MSE & MAE & MSE & MAE & MSE & MAE & MSE & MAE & MSE & MAE & MSE & MAE & MSE & MAE & MSE & MAE \\
\hline
ETTm1 & \secondres{0.054} & \boldres{0.147} & 0.120 & 0.253 & 0.104 & 0.218 & 0.093 & 0.206 & 0.062 & 0.177 & 0.717 & 0.570 & 0.071 & 0.188 & \boldres{0.050} & \secondres{0.154} & 0.989 & 0.786 & 0.516 & 0.497 & 0.113 & 0.254 \\
ETTm2 & \boldres{0.030} & \boldres{0.103} & 0.208 & 0.327 & \secondres{0.046} & \secondres{0.151} & 0.096 & 0.208 & 0.101 & 0.215 & 0.465 & 0.508 & 0.156 & 0.292 & 0.157 & 0.280 & 1.027 & 0.800 & 0.266 & 0.407 & 0.175 & 0.324 \\
ETTh1 & 0.129 & \boldres{0.235} & 0.202 & 0.329 & 0.284 & 0.373 & 0.201 & 0.306 & \boldres{0.117} & 0.246 & 0.842 & 0.682 & 0.161 & 0.279 & \secondres{0.122} & \secondres{0.245} & 1.225 & 0.873 & 0.621 & 0.571 & 0.424 & 0.481 \\
ETTh2 & \boldres{0.067} & \boldres{0.166} & 0.367 & 0.436 & \secondres{0.119} & \secondres{0.250} & 0.142 & 0.259 & 0.163 & 0.279 & 1.079 & 0.792 & 0.337 & 0.452 & 0.234 & 0.352 & 2.039 & 1.114 & 0.431 & 0.503 & 0.495 & 0.475 \\
Electricity & \boldres{0.086} & \boldres{0.202} & 0.214 & 0.339 & 0.131 & 0.262 & 0.132 & 0.260 & \secondres{0.130} & \secondres{0.259} & 0.297 & 0.382 & 0.222 & 0.328 & 0.200 & 0.313 & 0.277 & 0.365 & 0.582 & 0.597 & 0.222 & 0.293 \\
Weather & \boldres{0.036} & \boldres{0.070} & 0.076 & 0.171 & 0.055 & 0.117 & 0.052 & 0.110 & 0.099 & 0.203 & 0.152 & 0.235 & 0.045 & 0.104 & \secondres{0.038} & \secondres{0.087} & 0.365 & 0.434 & 0.183 & 0.291 & 0.045 & 0.108 \\
Exchange & \boldres{0.005} & \boldres{0.041} & 0.128 & 0.262 & 0.042 & 0.142 & 0.046 & 0.146 & 0.241 & 0.324 & 0.398 & 0.532 & 0.439 & 0.554 & 0.302 & 0.439 & \secondres{0.032} & \secondres{0.123} & 0.043 & 0.146 & 0.040 & 0.139 \\
\hline
1st Count & 5 & 6 & 0 & 0 & 0 & 0 & 0 & 0 & 1 & 0 & 0 & 0 & 0 & 0 & 1 & 0 & 0 & 0 & 0 & 0 & 0 & 0 \\
\hline
\end{tabular}
}
\end{table*}

%% file: tables/anomaly.tex
\begin{table*}[!ht]
\centering
\small
\caption{Anomaly detection results measured by F1-score (\%). A higher F1 indicates better performance. Full results are available in Table~\ref{tab:full_anomaly_detection_results}. \boldres{Red bold}: the best results; \secondres{blue}: the second best.}
\label{tab:anomaly}
\resizebox{\textwidth}{!}{
\begin{tabular}{l|cccccccccccc|c}
\toprule
\multirow{2}{*}{Datasets} & 
\rotatebox{0}{\scalebox{0.8}{Transformer}} & 
\rotatebox{0}{\scalebox{0.8}{Reformer}} & 
\rotatebox{0}{\scalebox{0.8}{Informer}} & 
\rotatebox{0}{\scalebox{0.8}{Autoformer}} & 
\rotatebox{0}{\scalebox{0.8}{Pyraformer}} & 
\rotatebox{0}{\scalebox{0.8}{Anomaly T.}} & 
\rotatebox{0}{\scalebox{0.8}{Stationary}} & 
\rotatebox{0}{\scalebox{0.8}{ETSformer}} & 
\rotatebox{0}{\scalebox{0.8}{DLinear}} & 
\rotatebox{0}{\scalebox{0.8}{TiDE}} & 
\rotatebox{0}{\scalebox{0.8}{iTransformer}} & 
\rotatebox{0}{\scalebox{0.8}{PatchTST}} & 
\rotatebox{0}{\scalebox{0.8}{\textbf{MAS4TS}}} \\
& \scalebox{0.7}{\citeyear{vaswani2017attention}} 
& \scalebox{0.7}{\citeyear{kitaev2020reformer}} 
& \scalebox{0.7}{\citeyear{zhou2021informer}} 
& \scalebox{0.7}{\citeyear{wu2021autoformer}} 
& \scalebox{0.7}{\citeyear{liu2022pyraformer}} 
& \scalebox{0.7}{\citeyear{xu2021anomaly}} 
& \scalebox{0.7}{\citeyear{liu2022non}} 
& \scalebox{0.7}{\citeyear{woo2022etsformer}} 
& \scalebox{0.7}{\citeyear{zeng2023dlinear}} 
& \scalebox{0.7}{\citeyear{das2023long}} 
& \scalebox{0.7}{\citeyear{liu2023itransformer}} 
& \scalebox{0.7}{\citeyear{nie2023patchtst}} 
& \scalebox{0.7}{(Ours)} \\
\midrule
SMD & 79.56 & 75.32 & 81.65 & 85.11 & 83.04 & 85.49 & 84.62 & 83.13 & 77.10 & 68.91 & 71.15 & 84.62 & 85.10 \\
MSL & 78.68 & 84.40 & 84.06 & 79.05 & 84.86 & 83.31 & 77.50 & 85.03 & 84.88 & 70.18 & 72.54 & 78.70 & 82.06 \\
SMAP & 69.70 & 70.40 & 69.92 & 71.12 & 71.09 & 71.18 & 71.09 & 69.50 & 69.26 & 64.00 & 66.87 & 68.82 & 68.58 \\
SWaT & 80.37 & 82.80 & 81.43 & 92.74 & 91.78 & 83.10 & 79.88 & 84.91 & 87.52 & 76.73 & 79.18 & 85.72 & 93.14 \\
PSM & 76.07 & 73.61 & 77.10 & 93.29 & 82.08 & 79.40 & 97.29 & 91.76 & 93.55 & 92.50 & 95.17 & 96.08 & 95.04 \\
\midrule
\textbf{Avg F1} & 76.88 & 77.31 & 78.83 & 84.26 & 82.57 & 80.50 & 82.08 & \secondres{82.87} & 82.46 & 74.46 & 76.98 & 82.79 & \boldres{84.78} \\
\bottomrule
\end{tabular}
}
\end{table*}

%% file: appendix/efficiency.tex
\subsection{Efficiency Study}
\label{sec:efficiency}

To evaluate the computational efficiency of \model, we compare it with representative pretrained LM-based time series methods under a unified experimental setting. All measurements are conducted on the forecasting task of the ETTh1 dataset with an input length of 96 and a prediction length of 96. We report the number of trainable parameters, peak GPU memory consumption, end-to-end training time per epoch, and inference latency per batch. \model uses an instruction-following VLM interface (\textit{Qwen3-VL-235B-Instruct} by default) by default for anchor extraction to balance output quality and efficiency. %All experiments are performed on a single NVIDIA RTX 4090 GPU with batch size 32.

\begin{table}[tbp]
\caption{Efficiency comparison between MAS4TS and pretrained LM-based time-series baselines on ETTh1 forecasting ($L{=}96$, $H{=}96$). We report trainable parameters, peak GPU memory, training time per epoch, and inference latency per batch. MAS4TS variants use \textit{Qwen3-VL-235B-Thinking} (Thinking) and \textit{Nano-Banana} (Generative), and the baselines use GPT-2 as the text encoding.}
  \label{tab:all_efficiency}
  \vskip 0.05in
  \centering
  \resizebox{0.95\columnwidth}{!}{
  \begin{threeparttable}
  \setlength{\tabcolsep}{4pt}
  \begin{tabular}{lcccc}
    \toprule
    Models & Params (M) & Mem (GB) & Train (s) & Infer (ms) \\
    \midrule
    \textbf{MAS4TS} & \textbf{5.38} & \textbf{0.05} & \textbf{24.55} & \textbf{90.49} \\
    \quad - Thinking & 5.38 & 0.05 & 35.98 & 133.75 \\
    \quad - Generative & 5.38 & 0.05 & 915.92 & 2767.01 \\
    \hline
    Time-VLM & 151.58 & 0.72 & 319.67 & 1067.14 \\
    UniTime$^\dagger$ & 107.32 & 5.18 & 53.49 & 165.77 \\
    TimeLLM$^\dagger$ & 134.90 & 15.96 & 853.75 & 2296.28 \\
    \bottomrule
  \end{tabular}
  \begin{tablenotes}
    \footnotesize
    \item[*] Experiments conducted on NVIDIA 4090 with batch size 32. 
    \item[*] The $\dagger$ represents the deployment and inference of local LMs, while we reason through the EAS API.
  \end{tablenotes}
  \end{threeparttable}
  }
  \vspace{-4mm}
\end{table}

Table~\ref{tab:all_efficiency} summarizes the efficiency comparison results. Although \model integrates a multi-agent workflow with foundation models, it maintains a compact backbone (5.38M parameters) and extremely low peak GPU memory footprint (0.05\,GB), which are orders of magnitude smaller than pretrained LM-based methods such as Time-VLM, UniTime, and TimeLLM. Concretely, these baselines involve over $100$M parameters and consume several to tens of gigabytes of GPU memory, whereas \model keeps the learnable core lightweight and invokes the VLM only when needed, in a mini-batched and parallel manner across samples. Default \textit{Instruct-style} anchoring provides the best efficiency-quality trade-off and extracts anchors using a structured, schema-constrained interface that avoids long, free-form decoding, while both thinking and generative variants introduce heavier reasoning overhead.
The results indicate that the practical efficiency of \model comes primarily from lightweight coordination plus structured tool usage, rather than scaling a monolithic architecture.

%% file: contents/05conclusion.tex
\section{Conclusion}
We presented MAS4TS, a tool-driven multi-agent framework for general time-series analysis built on an Analyzer--Reasoner--Executor paradigm. MAS4TS combines VLM-based visual anchoring with anchor-guided latent trajectory reconstruction. It further enables adaptive routing and tool-chain execution to solve forecasting, classification, imputation, and anomaly detection within a unified protocol.

%% file: appendix/A_Experimental_Details.tex
\section{Experiment Details}
\label{appx:experiment_details}

\subsection{Dataset Details}
\label{appx:dataset_details}

\input{tables/dataset-details}
As shown in Table \ref{tab:dataset}, the benchmark datasets span diverse domains and temporal granularities:
\begin{itemize}[leftmargin=*, itemsep=2pt]
    \item \textbf{Long-term Forecasting}: 7 datasets covering electricity transformer temperature (ETT series with 4 subsets), electricity consumption (Electricity), transportation flow (Traffic), meteorological indicators (Weather), solar power generation (Solar-Energy), and currency exchange rates (Exchange).
    \item \textbf{Imputation}: The same forecasting datasets with artificially induced random missing patterns at 12.5\%, 25\%, 37.5\%, and 50\% mask ratios, following the protocol established in \citep{wu2023timesnet}.
    \item \textbf{Classification}: 8 representative datasets from the UEA multivariate time series archive \citep{bagnall2018uea}, covering chemical sensing (EthanolConcentration), video analysis (FaceDetection), motion capture (Handwriting, UWaveGestureLibrary), medical diagnostics (Heartbeat), audio recognition (JapaneseVowels), and brain-computer interface (SelfRegulationSCP1, SelfRegulationSCP2).
    \item \textbf{Anomaly Detection}: 5 real-world datasets from server machines (SMD, PSM), spacecraft telemetry systems (MSL, SMAP), and water treatment infrastructure (SWaT).
\end{itemize}

\subsection{Baseline Methods}
\label{appx:baselines}
We reproduce all baselines using the official codebases and/or settings reported in the original papers.
To ensure a fair comparison, we keep the \emph{input embedding} and \emph{final projection} interfaces consistent across neural models when applicable, and focus on comparing their core sequence modeling or task-specific mechanisms.
Note that some baselines are only evaluated on specific tasks. %(see the task paragraphs below).

\paragraph{(1) Time-series specialized models.}
\begin{itemize}[leftmargin=*, itemsep=2pt]
\item \textbf{Transformer-based time-series models}:
\begin{itemize}[leftmargin=*, itemsep=1pt]
\item \textit{iTransformer} \citep{liu2023itransformer}: performs self-attention across variates (inverted tokenization) to better model multivariate dependencies.
\item \textit{PatchTST} \citep{nie2023patchtst}: uses patching and channel-independence to reduce sequence length and stabilize long-horizon learning.
\item \textit{Crossformer} \citep{zhang2023crossformer}: models cross-dimension dependencies via a two-stage attention design (intra-/inter-dimension).
\item \textit{Non-stationary Transformer} \citep{liu2022non} (reported as \textit{Stationary}): mitigates distribution shift with series stationarization and de-stationary attention.
\item \textit{Autoformer} \citep{wu2021autoformer}: replaces point-wise attention with auto-correlation and progressive series decomposition for long-term forecasting.
\item \textit{FEDformer} \citep{zhou2022fedformer}: conducts frequency-enhanced decomposition and attention in the spectral domain for efficient long-sequence modeling.
\item \textit{ETSformer} \citep{woo2022etsformer}: injects exponential smoothing (ETS-style) components to explicitly capture level/trend/seasonality.
\item \textit{Informer} \citep{zhou2021informer}: adopts ProbSparse attention to reduce attention cost for long sequence forecasting.
\item \textit{Reformer} \citep{kitaev2020reformer}: uses LSH attention to approximate softmax attention with lower memory/computation.
\item \textit{Pyraformer} \citep{liu2022pyraformer}: builds pyramidal attention across multiple temporal resolutions to model long-range structure.
\end{itemize}

\item \textbf{CNN/MLP-style time-series models}:
\begin{itemize}[leftmargin=*, itemsep=1pt]
\item \textit{TimesNet} \citep{wu2023timesnet}: reshapes 1D series into 2D representations guided by multi-periodicity and applies 2D convolution blocks.
\item \textit{DLinear} \citep{zeng2023dlinear}: decomposes series into trend and residual parts and forecasts with lightweight linear layers.
\item \textit{LightTS} \citep{campos2023lightts}: leverages efficient MLP mixing with interval/continuous sampling to handle long contexts.
\item \textit{TiDE} \citep{das2023long}: encodes the look-back with covariates via dense residual blocks and decodes future values with a per-time-step temporal decoder.
\item \textit{SCINet} \citep{liu2022scinet}: recursively splits and interacts subsequences to capture hierarchical temporal dependencies.
\item \textit{TCN} \citep{he2019temporal}: uses dilated causal convolutions to model long receptive fields without recurrence.
\end{itemize}

\item \textbf{RNN/State-space style models}:
\begin{itemize}[leftmargin=*, itemsep=1pt]
\item \textit{LSTM} \citep{hochreiter1997long}: recurrent gating mechanism for learning long/short-term temporal dependencies.
\item \textit{LSTNet} \citep{lai2018modeling}: combines CNN feature extraction with RNN temporal modeling and skip connections for periodic patterns.
\item \textit{LSSL} \citep{gu2022efficiently}: reparameterizes HiPPO-initialized state space models as diagonal-plus-low-rank, enabling near-linear-time long-sequence modeling via efficient Cauchy-kernel convolution..
\end{itemize}
\end{itemize}

\paragraph{(2) Pretrained LM-based baselines.}
\begin{itemize}[leftmargin=*, itemsep=2pt]
\item \textit{MAFS} \citep{huang2025many}: employs a multi-agent framework where each agent specializes in a forecasting sub-task, with inter-agent communication and a voting aggregator for final prediction. (Note: we include it here as it shares the pre-train then fine-tune paradigm, though it pre-trains iTransformer on sub-tasks rather than leveraging large language models.)
\item \textit{Time-VLM} \citep{zhong2025time}: integrates temporal, visual, and textual modalities through pretrained Vision-Language Models (e.g., ViLT, CLIP). Time series are transformed into images via frequency/periodicity encoding and multi-scale convolution, while contextual text prompts are generated from statistical features. A frozen VLM encodes both modalities, and cross-modal attention fuses them with temporal memory features for forecasting.
\item \textit{UniTime} \citep{liu2024unitime}: employs pretrained language models with domain instructions. It adopts channel-independence and patch-based tokenization, uses natural language instructions to provide domain identification and alleviate domain confusion, and applies masking to balance convergence speeds across domains. The Language-TS Transformer aligns time series from various input spaces to the common latent space of language models for cross-domain generalization.
\end{itemize}

\paragraph{(3) Classical baselines and others.}
\begin{itemize}[leftmargin=*, itemsep=1pt]
\item\textit{Transformer}\citep{vaswani2017attention}: the vanilla encoder--decoder attention architecture as a general-purpose sequence baseline.
\item \textit{DTW} \citep{Berndt1994UsingDT}: uses Dynamic Time Warping distance (typically with 1-NN) as a strong classical baseline for time-series classification.
\item \textit{XGBoost} \citep{Chen2016XGBoostAS}: gradient-boosted trees trained on engineered temporal features/statistics.
\end{itemize}
For anomaly detection, we additionally compare against \textit{Anomaly Transformer} \citep{xu2021anomaly}, a method specifically designed for this task that identifies anomalies via association discrepancy between series-wise and prior associations. Beyond task-specific models, we also consider classical methods that remain highly competitive on experimental results. For instance, in classification, \textit{DTW}, \textit{XGBoost}, and the vanilla \textit{Transformer} have all demonstrated strong performance and are thus included as additional baselines.

\subsection{Evaluation Metrics}

For the metrics, we adopt different evaluation criteria tailored to each task:

\paragraph{Long-term Forecasting and Imputation.} We use mean square error (MSE) and mean absolute error (MAE):
\begin{align*}
    \text{MSE} &= \frac{1}{H \cdot C} \sum_{i=1}^H \sum_{j=1}^C (\mathbf{X}_{i,j} - \widehat{\mathbf{X}}_{i,j})^2, &
    \text{MAE} &= \frac{1}{H \cdot C} \sum_{i=1}^H \sum_{j=1}^C |\mathbf{X}_{i,j} - \widehat{\mathbf{X}}_{i,j}|,
\end{align*}
where $\mathbf{X},\widehat{\mathbf{X}}\in\mathbb{R}^{H\times C}$ denote the ground truth and predicted values with $H$ time steps and $C$ feature dimensions. Lower values indicate better performance.

\paragraph{Classification.} We adopt accuracy (ACC) as the primary evaluation metric:
\begin{align*}
    \text{ACC} = \frac{\text{Number of Correct Predictions}}{\text{Total Number of Samples}} \times 100\%.
\end{align*}
Higher accuracy indicates better classification performance.

\paragraph{Anomaly Detection.} We adopt precision (P), recall (R), and F1-score (F1) following the point-adjust protocol \citep{xu2021anomaly}:
\begin{align*}
    \text{Precision} &= \frac{\text{TP}}{\text{TP} + \text{FP}}, &
    \text{Recall} &= \frac{\text{TP}}{\text{TP} + \text{FN}}, &
    \text{F1} &= 2 \cdot \frac{\text{Precision} \cdot \text{Recall}}{\text{Precision} + \text{Recall}},
\end{align*}
where TP, FP, and FN denote true positives, false positives, and false negatives, respectively. The F1-score provides a balanced measure between precision and recall, with higher values indicating better detection performance.

\subsection{Model and Agent Configuration}
\label{appx:model_agent_config}

This section provides a comprehensive overview of the model hyperparameters, agent-specific configurations, and shared memory mechanism in \model. The framework adopts a modular design where each agent maintains specialized parameter sets while coordinating through a unified memory interface.

\paragraph{Training Configuration.}
Table~\ref{tab:model_config} summarizes the task-specific training configurations. All experiments employ the AdamW optimizer with default momentum parameters $(\beta_1, \beta_2) = (0.9, 0.999)$ and weight decay of $10^{-2}$. We apply early stopping with a patience of 3 epochs to prevent overfitting and use cosine annealing for learning rate scheduling. Mixed-precision training (FP16) is enabled by default to accelerate computation while maintaining numerical stability.

\begin{table}[thbp]
  \vspace{-5pt}
  \caption{Task-specific training configurations for \model. The embedding dimension $d_{\text{model}}$ and network depth are kept consistent across tasks to ensure fair comparison, while learning rates and batch sizes are tuned per task for optimal convergence.}\label{tab:model_config}
  \vskip 0.05in
  \centering
  \begin{threeparttable}
  \begin{small}
  \renewcommand{\multirowsetup}{\centering}
  \setlength{\tabcolsep}{3.8pt}
  \begin{tabular}{l|cccc|cccc}
    \toprule
    \multirow{2}{*}{\textbf{Task}} & \multicolumn{4}{c|}{\textbf{Architecture}} & \multicolumn{4}{c}{\textbf{Training}} \\
    \cmidrule(lr){2-5}\cmidrule(lr){6-9}
    & $d_{\text{model}}$ & Layers & Patch & Stride & LR & Loss & Batch & Epochs\\
    \midrule
    Long-term Forecast & 64 & 2 & 16 & 8 & $10^{-4}$ & MSE & 32 & 10 \\
    Imputation & 64 & 2 & 16 & 8 & $10^{-3}$ & MSE & 16 & 10 \\
    Classification & 64 & 2 & 16 & 8 & $10^{-3}$ & CE & 16 & 30 \\
    Anomaly Detection & 64 & 3 & 16 & 8 & $10^{-4}$ & MSE & 128 & 10 \\
    \bottomrule
  \end{tabular}
  \begin{tablenotes}
    \footnotesize
    \item[] LR: initial learning rate with cosine annealing decay. CE: Cross-Entropy loss.
  \end{tablenotes}
  \end{small}
  \end{threeparttable}
  \vspace{-5pt}
\end{table}

\paragraph{Ablation Flags.}
\model supports fine-grained ablation studies through configurable flags that enable/disable individual components:
\begin{itemize}[leftmargin=*, itemsep=1pt]
    \item \texttt{enable\_visual\_reasoner}: Toggle VLM-based anchor extraction (default: True)
    \item \texttt{enable\_numeric\_reasoner}: Toggle Neural ODE sequence reconstruction (default: True)
    \item \texttt{enable\_shared\_memory}: Toggle cross-agent memory sharing (default: True)
    \item \texttt{enable\_gated\_attention}: Toggle gated vs. simple MLP communication (default: True)
    \item \texttt{enable\_tools}: Toggle tool-augmented execution vs. simple MLP heads (default: True)
    \item \texttt{completion\_strategy}: ODE solver alternative: \{`ode', `linear', `quadratic', `repeat'\}
\end{itemize}

\input{tables/model-parameters}

\subsection{Implementation Details}
We provide the dataset descriptions and experiment configurations in Table~\ref{tab:dataset} and~\ref{tab:model_config}. All experiments are conducted on a single NVIDIA GeForce RTX 4090 GPU (48GB version) with a fixed random seed for reproducibility. The MAS4TS framework is implemented in PyTorch 2.9.1+cu128 with CUDA 12.8. For the Vision-Language Model component, we utilize the \textit{Qwen2-VL-235B-Instruct} model accessed through EAS API services of the Alibaba Cloud Platform. 

%% file: tables/dataset-details.tex
\begin{table}[thbp]
  \caption{Dataset descriptions. The dataset size is organized in (Train, Validation, Test).}\label{tab:dataset}
  \vskip 0.05in
  \centering
  \begin{threeparttable}
  \begin{small}
  \renewcommand{\multirowsetup}{\centering}
  \setlength{\tabcolsep}{3.8pt}
  \begin{tabular}{c|l|c|c|c|c}
    \toprule
    Tasks & Dataset & Dim & Series Length & Dataset Size & \scalebox{0.8}{Information (Frequency)} \\
    \toprule
     & ETTm1, ETTm2 & 7 & \scalebox{0.8}{\{96, 192, 336, 720\}} & (34465, 11521, 11521) & \scalebox{0.8}{Electricity (15 mins)}\\
    \cmidrule{2-6}
     & ETTh1, ETTh2 & 7 & \scalebox{0.8}{\{96, 192, 336, 720\}} & (8545, 2881, 2881) & \scalebox{0.8}{Electricity (Hourly)} \\
    \cmidrule{2-6}
    Long-term & Weather & 21 & \scalebox{0.8}{\{96, 192, 336, 720\}} & (36792, 5271, 10540) & \scalebox{0.8}{Weather (10 mins)} \\
    \cmidrule{2-6}
    Forecasting & Solar-Energy & 137 & \scalebox{0.8}{\{96, 192, 336, 720\}} & (36345, 5065, 10321) & \scalebox{0.8}{Solar (10 mins)} \\
    \cmidrule{2-6}
     & Exchange & 8 & \scalebox{0.8}{\{96, 192, 336, 720\}} & (5120, 665, 1422) & \scalebox{0.8}{Exchange rate (Daily)}\\
    \midrule
     & ETTm1, ETTm2 & 7 & 96 & (34465, 11521, 11521) & \scalebox{0.8}{Electricity (15 mins)} \\
    \cmidrule{2-6}
     & ETTh1, ETTh2 & 7 & 96 & (8545, 2881, 2881) & \scalebox{0.8}{Electricity (Hourly)}\\
    \cmidrule{2-6}
    Imputation & Electricity & 321 & 96 & (18317, 2633, 5261) & \scalebox{0.8}{Electricity (Hourly)}\\
    \cmidrule{2-6}
     & Weather & 21 & 96 & (36792, 5271, 10540) & \scalebox{0.8}{Weather (10 mins)} \\
    \cmidrule{2-6}
    & Exchange & 8 & 96 & (5120, 665, 1422) & \scalebox{0.8}{Exchange rate (Daily)}\\
    \midrule
     & \scalebox{0.8}{EthanolConcentration} & 3 & 1751 & (261, 0, 263) & \scalebox{0.8}{Alcohol Industry}\\
    \cmidrule{2-6}
    & \scalebox{0.8}{FaceDetection} & 144 & 62 & (5890, 0, 3524) & \scalebox{0.8}{Face (250Hz)}\\
    \cmidrule{2-6}
    & \scalebox{0.8}{Handwriting} & 3 & 152 & (150, 0, 850) & \scalebox{0.8}{Handwriting}\\
    \cmidrule{2-6}
    Classification & \scalebox{0.8}{Heartbeat} & 61 & 405 & (204, 0, 205)& \scalebox{0.8}{Heart Beat}\\
    \cmidrule{2-6}
    (UEA) & \scalebox{0.8}{JapaneseVowels} & 12 & 29 & (270, 0, 370) & \scalebox{0.8}{Voice}\\
    \cmidrule{2-6}
    & \scalebox{0.8}{SelfRegulationSCP1} & 6 & 896 & (268, 0, 293) & \scalebox{0.8}{Health (256Hz)}\\
    \cmidrule{2-6}
    & \scalebox{0.8}{SelfRegulationSCP2} & 7 & 1152 & (200, 0, 180) & \scalebox{0.8}{Health (256Hz)}\\
    \cmidrule{2-6}
    & \scalebox{0.8}{UWaveGestureLibrary} & 3 & 315 & (120, 0, 320) & \scalebox{0.8}{Gesture}\\
    \midrule
     & SMD & 38 & 100 & (566724, 141681, 708420) & \scalebox{0.8}{Server Machine} \\
    \cmidrule{2-6}
    Anomaly & MSL & 55 & 100 & (44653, 11664, 73729) & \scalebox{0.8}{Spacecraft} \\
    \cmidrule{2-6}
    Detection & SMAP & 25 & 100 & (108146, 27037, 427617) & \scalebox{0.8}{Spacecraft} \\
    \cmidrule{2-6}
     & SWaT & 51 & 100 & (396000, 99000, 449919) & \scalebox{0.8}{Infrastructure} \\
    \cmidrule{2-6}
    & PSM & 25 & 100 & (105984, 26497, 87841)& \scalebox{0.8}{Server Machine} \\
    \bottomrule
    \end{tabular}
    \end{small}
  \end{threeparttable}
\end{table}

%% file: tables/model-parameters.tex
\begin{table}[htbp]
  \centering
  \caption{MAS4TS Model Architecture Parameters}
  \label{tab:model_params}
  \begin{small}
    \scalebox{0.95}{
      \begin{tabular}{l|c|p{5.8cm}}
        \toprule
        \textbf{Parameter} & \textbf{Value} & \textbf{Description} \\
        \midrule
        \multicolumn{3}{l}{\textit{Global Configuration}} \\
        \midrule
        d\_model & 64 & Embedding dimension for all agents \\
        d\_memory & 64 & Shared memory dimension \\
        patch\_len & 16 & Patch length for tokenization \\
        stride & 8 & Stride between patches \\
        \midrule
        \multicolumn{3}{l}{\textit{Data Analyzer Agent}} \\
        \midrule
        top\_k\_features & 10 & Top-k features for covariance selection \\
        feature\_method & covariance & Feature selection criterion \\
        \midrule
        \multicolumn{3}{l}{\textit{Visual Reasoner Agent}} \\
        \midrule
        vlm\_model & Qwen3-VL-235B-Instruct & Vision-Language Model backbone \\
        max\_anchors & 20 & Maximum anchor points per sample \\
        confidence\_threshold & 0.7 & Minimum confidence for anchor acceptance \\
        \midrule
        \multicolumn{3}{l}{\textit{Numeric Reasoner Agent}} \\
        \midrule
        hidden\_dim & 128 & Hidden dimension for ODE dynamics \\
        ode\_solver & RK4 & ODE integration method \\
        ode\_step & 0.05 & Integration step size \\
        \midrule
        \multicolumn{3}{l}{\textit{Task Executor Agent}} \\
        \midrule
        e\_layers & 2 & Encoder layers in task tools \\
        n\_heads & 4 & Attention heads \\
        dropout & 0.1 & Dropout rate \\
        \bottomrule
      \end{tabular}
    }
  \end{small}
\end{table}

%% file: appendix/B_Complete_Results.tex
\section{Complete Results}
\label{appx:complete_results}

This appendix provides comprehensive experimental results for all four evaluation tasks: long-term forecasting, time series classification, imputation, and anomaly detection.

\subsection{Long-term Forecasting}
\label{appx:forecasting_results}

\input{tables/forecasting_full}

Table~\ref{tab:full_long_term_forecasting_results} presents complete forecasting results across all datasets and prediction horizons $H \in \{96, 192, 336, 720\}$. The input sequence length is fixed at $L = 96$ for all methods following prior work~\citep{wu2022timesnet}. MAS4TS consistently achieves competitive or superior performance compared to both task-specific models (e.g., PatchTST, iTransformer) and pretrained LM-based methods (e.g., UniTime, TimeVLM).

\subsection{Classification}
\label{appx:classification_results}
\input{tables/classification_full}

Table~\ref{tab:full_classification_results} shows classification accuracy on 8 representative datasets from the UEA multivariate time series archive~\citep{bagnall2018uea}. These datasets span diverse application domains including chemical sensing, video analysis, motion capture, medical diagnostics, and audio recognition.

\subsection{Imputation}
\label{appx:imputation_results}
\input{tables/imputation_full}

Table~\ref{tab:full_imputation_results} presents imputation results under different missing rates $r \in \{12.5\%, 25\%, 37.5\%, 50\%\}$. Following the experimental protocol in TimesNet~\citep{wu2023timesnet}, we randomly mask values in the test sequences and evaluate reconstruction quality using MSE and MAE metrics.

\subsection{Anomaly Detection}
\label{appx:anomaly_results}
\input{tables/anomaly_full}

Table~\ref{tab:full_anomaly_detection_results} reports anomaly detection results on five benchmark datasets using precision (P), recall (R), and F1-score (F1) metrics. Following standard practice~\citep{xu2021anomaly}, we apply the point-adjustment protocol where a detection is considered correct if any point within an anomalous segment is identified.

%% file: tables/forecasting_full.tex
\begin{table}[htbp]
  \caption{Full results for the long-term forecasting task. We compare extensive competitive models under different prediction lengths. The input sequence length is set to 96 for all datasets. \emph{Avg} is averaged from all four prediction lengths, that $\{96, 192, 336, 720\} $. \boldres{Red bold}: the best results; \secondres{underlined in blue}: the second best.}\label{tab:full_long_term_forecasting_results}
  \vskip 0.05in
  \centering
  \resizebox{1.0\columnwidth}{!}{
  \begin{threeparttable}
  \begin{small}
  \renewcommand{\multirowsetup}{\centering}
  \setlength{\tabcolsep}{1.2pt}
  \begin{tabular}{c|c|cc|cc|cc|cc|cc|cc|cc|cc|cc|cc|cc|cc|cc|cc}
    \toprule
    \multicolumn{2}{c}{\multirow{2}{*}{Models}} & \multicolumn{2}{c}{\rotatebox{0}{{\textbf{MAS4TS}}}} &
    \multicolumn{2}{c}{\rotatebox{0}{{MAFS}}} &
    \multicolumn{2}{c}{\rotatebox{0}{{TimeVLM}}} &
    \multicolumn{2}{c}{\rotatebox{0}{{UniTime}}} &
    \multicolumn{2}{c}{\rotatebox{0}{{iTransformer}}} &
    \multicolumn{2}{c}{\rotatebox{0}{{PatchTST}}} &
    \multicolumn{2}{c}{\rotatebox{0}{{Crossformer}}} &
    \multicolumn{2}{c}{\rotatebox{0}{{TiDE}}} &
    \multicolumn{2}{c}{\rotatebox{0}{{TimesNet}}} & 
    \multicolumn{2}{c}{\rotatebox{0}{{DLinear}}} &
    \multicolumn{2}{c}{\rotatebox{0}{{SCINet}}}&
    \multicolumn{2}{c}{\rotatebox{0}{{FEDformer}}} & 
    \multicolumn{2}{c}{\rotatebox{0}{{Stationary}}} & 
    \multicolumn{2}{c}{\rotatebox{0}{{Autoformer}}}  \\
    \multicolumn{2}{c}{} & \multicolumn{2}{c}{{(\textbf{Ours})}} &
    \multicolumn{2}{c}{{\citeyear{huang2025many}}} &
    \multicolumn{2}{c}{{\citeyear{zhong2025time}}} &
    \multicolumn{2}{c}{{\citeyear{liu2024unitime}}} &
    \multicolumn{2}{c}{{\citeyear{liu2023itransformer}}} &
    \multicolumn{2}{c}{{\citeyear{nie2023patchtst}}}&
    \multicolumn{2}{c}{{\citeyear{zhang2023crossformer}}}&
    \multicolumn{2}{c}{{\citeyear{das2023long}}}&
    \multicolumn{2}{c}{{\citeyear{wu2022timesnet}}}&
    \multicolumn{2}{c}{{\citeyear{zeng2023dlinear}}}&
    \multicolumn{2}{c}{{\citeyear{liu2022scinet}}}&
    \multicolumn{2}{c}{{\citeyear{zhou2022fedformer}}}&
    \multicolumn{2}{c}{{\citeyear{liu2022non}}}&
    \multicolumn{2}{c}{{\citeyear{wu2021autoformer}}}
    \\
    \hline    
    % \cmidrule(lr){3-4} \cmidrule(lr){5-6}\cmidrule(lr){7-8} \cmidrule(lr){9-10}\cmidrule(lr){11-12}\cmidrule(lr){13-14}\cmidrule(lr){15-16}\cmidrule(lr){17-18}\cmidrule(lr){19-20} \cmidrule(lr){21-22} \cmidrule(lr){23-24} \cmidrule(lr){25-26}
    \multicolumn{2}{c|}{Metric} & {MSE} & {MAE} & {MSE} & 
    {MAE} & {MSE} & 
    {MAE} & {MSE} & 
    {MAE} & {MSE} & 
    {MAE} & {MSE} & {MAE} & {MSE} & {MAE} & {MSE} & {MAE} & {MSE} & {MAE} & {MSE} & {MAE} & {MSE} & {MAE} & {MSE} & {MAE} & {MSE} & {MAE}& {MSE} & {MAE}\\
    \toprule

 \multirow{5}{*}{\rotatebox{90}{{Weather}}} 
    & {96} &\boldres{{0.147}}&\boldres{{0.197}}& \secondres{0.166} & \secondres{0.208} & 0.181 & 0.220 & 0.171 & 0.214 & {0.174} & 0.214&{{0.186}}&{{0.227}}&{0.195}&{0.271}& {0.202} & {0.261} & 0.172 &{{0.220}}&{0.195}&{0.252} & {0.221} &{0.306}& {0.217} &{0.296}& {0.173} &{0.223} & {0.266} &{0.336}  \\
    & {192} &\boldres{{0.192}}&\boldres{{0.244}} & 0.218 & \secondres{0.253} & 0.227 & 0.260 & 0.217 & 0.254 & {0.221} & 0.254 & {0.234}&{{0.265}}&\secondres{{0.209}}&{0.277}& {0.242} & {0.298} &{{0.219}} &{{0.261}}&{0.237}&{0.295} & {0.261} &{0.340}& {0.276} &{0.336} & {0.245} &{0.285} & {0.307} &{0.367} \\
    & {336} &\boldres{{0.245}}&\boldres{{0.286}}& 0.274 & 0.295 & 0.281 & 0.298 & 0.274 & \secondres{0.293} & {0.278} & 0.296&{{0.284}}&{0.301}&\secondres{{0.273}}&{0.332}& {0.287} & {0.335} &{{0.280}} &{{0.306}}&{0.282}&{0.331}  & {0.309} &{0.378}& {0.339} &{0.380} & {0.321} &{0.338} & {0.359} &{0.395}\\
    & {720} &\boldres{{0.328}}&\boldres{{0.339}}&0.351 & 0.345 & 0.358 & 0.348 & 0.351 & \secondres{0.343} & {0.358} & 0.347&{0.356}&{0.349}&{0.379}&{0.401}& {{0.351}} & {0.386} &{0.365} &{{0.359}}&\secondres{{0.345}}&{0.382} & {0.377} &{0.427} & {0.403} &{0.428} & {0.414} &{0.410} & {0.419} &{0.428} 
    \\
    \cmidrule(lr){2-30}
    & {Avg} &\boldres{{0.228}}&\boldres{{0.267}}& \secondres{0.252} & \secondres{0.275} & 0.262 & 0.281 & 0.253 & 0.276 & 0.258 & 0.278&{{0.265}}&{0.285}&{0.264}&{0.320}& {0.271} & {0.320} &{{0.259}} &{{0.287}}&{0.265}&{0.315}&{0.292} &{0.363} &{0.309} &{0.360} &{0.288} &{0.314} &{0.338} &{0.382}  \\
    \midrule
    
    \multirow{5}{*}{\rotatebox{90}{{Solar-Energy}}} 
    & {96} &\boldres{{0.192}}& 0.259 & \secondres{0.198} & \boldres{0.237} & 0.220 & 0.269 & 0.239 & 0.301 &0.203 &\secondres{0.237}&{0.265}&{0.323}&{{0.232}}&{0.302} &{0.312} &{0.399}&{0.373}&{0.358}&{0.290}&{0.378}&{0.237}&{0.344}&{0.286}&{0.341}&{0.321}&{0.380}&{0.456}&{0.446}
    \\
    & {192} &\boldres{{0.207}}&0.271& \secondres{0.231} & \secondres{0.263} & 0.257 & 0.293 & 0.269 & 0.316 &0.233 &\boldres{{0.261}}&{0.288}&{0.332}&{0.371}&{0.410} &{0.339} &{0.416}&{0.397}&{0.376}&{0.320}&{0.398}&{0.280}&{0.380}&{0.291}&{0.337}&{0.346}&{0.369}&{0.588}&{0.561}
    \\
    & {336} &\boldres{{0.219}}&0.283& 0.250 & \secondres{0.278} & 0.271 & 0.303 & 0.288 & 0.324 &\secondres{{0.248}} &\boldres{{0.273}}&{0.301}&{0.339}&{0.495}&{0.515}&{0.368} &{0.430}&{0.420}&{0.380}&{0.353}&{0.415}&{0.304}&{0.389}&{0.354}&{0.416}&{0.357}&{0.387}&{0.595}&{0.588}\\
    & {720} &\boldres{{0.231}}&\secondres{{0.293}}& 0.255 & 0.281 & 0.287 & 0.307 & 0.285 & 0.320 & \secondres{{0.249}} &\boldres{{0.275}}&{0.295}&{0.336}&{0.526}&{0.542}&{0.370} &{0.425}&{0.420}&{0.381}&{0.357}&{0.413}&{0.308}&{0.388}
    &{0.380}&{0.437}&{0.375}&{0.424}&{0.733}&{0.633}
    \\
    \cmidrule(lr){2-30}
    & {Avg}&\boldres{{0.212}} &0.276 & 0.234 & \secondres{0.265} & 0.259 & 0.293 & 0.270 & 0.315 &\secondres{{0.233}} &\boldres{{0.262}}&{0.287}&{0.333}&{0.406}&{0.442}&{0.347} &{0.417}&{0.403}&{0.374}&{0.330}&{0.401}&{0.282} &{0.375}&{0.328} &{0.383} &{0.350} &{0.390} &{0.586} &{0.557} \\
    \midrule

    \multirow{5}{*}{\rotatebox{90}{{Exchange}}} 
    & {96}  &0.088&0.208& \secondres{0.084}	& \secondres{0.204} & \boldres{0.082} & \boldres{0.198} & 0.125 & 0.245& 0.086 & 0.206 & 0.088 &0.205 & {0.256} & {0.367} & {0.094} & {0.218} & {0.107} & {0.234} & {0.088} & {0.218} & {0.267} & {0.396} & {0.148} & {0.278} & {0.111} & {0.237} & {0.197} & {0.323} 
    \\
    & {192}  &0.181&{{0.303}} & 0.180 & 0.304 & \boldres{0.175} & \boldres{0.296} & 0.236 & 0.343 & {0.177} & \secondres{{0.299}} & \secondres{0.176} & \secondres{{0.299}} & {0.470} & {0.509} & {0.184} & {0.307} & {0.226} & {0.344} & 0.176 & {0.315} & {0.351} & {0.459} & {0.271} & {0.315} & {0.219} & {0.335} & {0.300} & {0.369} 
    \\
    & {336}  &\boldres{{0.293}}&\boldres{{0.396}} & 0.316 & 0.406 & 0.324 & 0.412 & 0.420 & 0.470 & {0.331} & \secondres{{0.417}}& \secondres{{0.301}} & \secondres{{0.397}} & {1.268} & {0.883} & {0.349} & {0.431} & {0.367} & {0.448} & {0.313} & {0.427} & {1.324} & {0.853} & {0.460} & {0.427} & {0.421} & {0.476} & {0.509} & {0.524}
    \\
    & {720}  &\boldres{{0.810}}&\boldres{{0.682}} & 1.002 & 0.755 & 0.840 & \secondres{0.689} & 1.078 & 0.791 &{{0.847}} & 0.691 & {0.901} & {0.714} & {1.767} & {1.068} & {0.852} & {0.698} & {0.964} & {0.746} & \secondres{{0.839}} & {{0.695}} & {1.058} & {0.797} & {1.195} & {{0.695}} & {1.092} & {0.769} & {1.447} & {0.941}
    \\
    \cmidrule(lr){2-30}
    & {Avg} & \boldres{{0.343}} & \boldres{0.397} & 0.396 & 0.417 & 0.355 & \secondres{0.399} & 0.465 & 0.462 &{{0.360}} & 0.403 & {0.367} & 0.404 & {0.940} & {0.707} & {0.370} & {0.413} & {0.416} & {0.443} & \secondres{{0.354}} & {0.414} & {0.750} & {0.626} & {0.519} & {0.429} & {0.461} & {0.454} & {0.613} & {0.539} \\
    \midrule

    \multirow{5}{*}{\rotatebox{90}{{ETTh1}}} 
    & {96} &\secondres{{0.378}}&\secondres{{0.399}}& 0.389 & 0.403 & \boldres{0.376} & \boldres{0.395} & 0.397 & 0.418 & 0.386 & \secondres{{0.405}}&{{0.460}}&{{0.447}}&{0.423}&{0.448} & {0.479}& {0.464}  &{{0.384}} &{{0.402}}&{0.397}&{0.412}&{0.654} &{0.599} 
     &{0.395} &{0.424} &{0.513} &{0.491} &{0.449} &{0.459} \\
    & {192} &\boldres{{0.419}}&\secondres{{0.428}}& 0.450 & 0.438 & \secondres{0.426} & \boldres{0.424} & 0.434 & 0.439 & {0.441} & {0.512}&{{0.477}}&0.429&{0.471}&{0.474}& {0.525} & {0.492} &0.436 &{{0.429}}&{0.446}&{0.441} &{0.719} &{0.631}
     &{0.469} &{0.470} &{0.534} &{0.504} &{0.500} &{0.482} 
    \\
    & {336} &\boldres{{0.459}}&\boldres{0.445} & 0.475 & \secondres{0.447} & 0.469 & 0.449 & \secondres{0.468} & 0.457 & 0.487 & 0.458&{0.546}&{0.496}&{0.570}&{0.546}& {0.565} & {0.515} &{0.491} &{0.469}&{{0.489}}&{{0.467}}&{0.778} &{0.659}
    &{0.530} &{0.499} &{0.588} &{0.535} &{0.521} &{0.496} 
    \\
    & {720} & 0.503 & {{0.501}} & \secondres{0.487}	& \boldres{0.472} & 0.559 & 0.504 & \boldres{0.469} & \secondres{0.477} & 0.503 & 0.491&{0.544}&{{0.517}}&{0.653}&{0.621}& {0.594} & {0.558} &{0.521} &0.500&{{0.513}}&{0.510}&{0.836} &{0.699}
    &{0.598} &{0.544} &{0.643} &{0.616} &{{0.514}} &{0.512}  
    \\
    \cmidrule(lr){2-30}
    & {Avg} &\boldres{{0.440}}&\boldres{{0.438}}& 0.450 & \secondres{0.440} & 0.457 & 0.443 & 0.442 & 0.448 & 0.454 & 0.447&{0.516}&{{0.484}}&{0.529}&{0.522}& {0.541} & {0.507} &{0.458} &{{0.450}}&{{0.461}}&{0.457}&{0.747} &{0.647}&{0.498} &{0.484} &{0.570} &{0.537} &{0.496} &{0.487} \\
    \midrule

    \multirow{5}{*}{\rotatebox{90}{{ETTh2}}} 
    & {96}&\secondres{{0.296}}&\secondres{{0.345}}& 0.298 & 0.346 & \boldres{0.290} & \boldres{0.339} & 0.296 & 0.345 & 0.297 & 0.349 &{0.308}&{0.355}&{0.745}&{0.584}&{0.400} & {0.440}  & {{0.340}} & {{0.374}}&{0.340}&{0.394}&{0.707} &{0.621}  &{0.358} &{0.397} &{0.476} &{0.458} &{0.346} &{0.388}  \\
    & {192} &0.386&0.399& 0.380 & 0.398 & \boldres{0.374} & \boldres{0.389} & \secondres{0.374} & \secondres{0.394} & 0.380 & 0.400&{0.393}&{0.405}&{0.877}&{0.656}& {0.528} & {0.509} & {{0.402}} & {{0.414}}&{0.482}&{0.479}&{0.860} &{0.689} &{0.429} &{0.439} &{0.512} &{0.493} &{0.456} &{0.452}  \\
    & {336} &\boldres{{0.407}}&\boldres{{0.422}} & \secondres{0.413} & \secondres{0.427} & 0.433 & 0.440 & 0.415 & 0.427 & {{0.428}} & 0.432&0.427 &{0.436}&{1.043}&{0.731}& {0.643} & {0.571}  & {{0.452}} & {{0.452}}&{0.591}&{0.541} &{1.000} &{0.744}&{0.496} &{0.487} &{0.552} &{0.551} &{0.482} &{0.486} \\
    & {720} &\secondres{{0.427}}&\boldres{{0.442}}& 0.428 & \secondres{0.444} & 0.438 & 0.452 & \boldres{0.425} & 0.444 & 0.427 & 0.445&{0.436}&{0.450}&{1.104}&{0.763}& {0.874} & {0.679} & {{0.462}} & {{0.468}}&{0.839}&{0.661} &{1.249} &{0.838}&{0.463} &{0.474}&{0.562} &{0.560} &{0.515} &{0.511}  
    \\
    \cmidrule(lr){2-30}
    & {Avg} &\boldres{{0.379}}&\boldres{{0.402}} & 0.380 & \secondres{0.403} & 0.384 & 0.405 & \secondres{0.379} & 0.403 & 0.383 & 0.407&{0.391}&{0.411}&{0.942}&{0.684}& {0.611} & {0.550}  &{{0.414}} &{{0.427}}&{0.563}&{0.519} &{0.954} &{0.723}&{0.437} &{0.449} &{0.526} &{0.516} &{0.450} &{0.459} \\
    \midrule

    \multirow{5}{*}{\rotatebox{90}{{ETTm1}}} 
    & {96} &\boldres{{0.296}}&\boldres{{0.341}}& 0.329 & \secondres{0.362} & \secondres{0.326} & 0.365 & 0.322 & 0.363 & 0.334 & 0.368&{{0.352}}&{0.374}&{0.404}&{0.426}& {0.364} & {0.387} &{{0.338}} &{{0.375}}&{0.346}&{0.374} &{0.418} &{0.438} &{0.379} &{0.419} &{0.386} &{0.398} &{0.505} &{0.475} 
    \\
    & {192} &\boldres{{0.341}}&\boldres{{0.366}}&0.370 & 0.385 & 0.368 & \secondres{0.384} & \secondres{0.366} & 0.387 & {0.390} & {0.393}&0.374&0.387&{0.450}&{0.451}&{0.398} & {0.404} &{{0.374}} &{{0.387}}&{0.382}&{0.391} &{0.439} &{0.450}&{0.426} &{0.441} &{0.459} &{0.444} &{0.553} &{0.496} 
    \\
    & {336} &\boldres{{0.385}}&\boldres{{0.390}}& 0.409 & 0.410 & \secondres{0.394} & \secondres{0.405} & 0.398 & 0.407 & {0.426} & {0.420}&{{0.421}}&{0.414}&{0.532}&{0.515}& {0.428} & {0.425} &0.410 &0.411&{0.415}&{0.415} &{0.490} &{0.485}&{0.445} &{0.459} &{0.495} &{0.464} &{0.621} &{0.537}  
    \\
    & {720} &\boldres{{0.445}}&\boldres{{0.426}}&0.478 & 0.451 & 0.462 & \secondres{0.440} & \secondres{0.454} & 0.440 & {0.491} & {0.459}&0.462&0.449&{0.666}&{0.589}& {0.487} & {0.461} &{{0.478}} &{0.450}&{0.473}&{0.451} &{0.595} &{0.550}&{0.543} &{0.490} &{0.585} &{0.516} &{0.671} &{0.561} 
    \\
    \cmidrule(lr){2-30}
    & {Avg} &\boldres{{0.367}}&\boldres{{0.381}}& 0.397 & 0.402 & 0.388 & \secondres{0.398} & \secondres{0.385} & 0.399 & {0.407} & {0.410}&{{0.406}}&{0.407}&{0.513}&{0.495}& {0.419} & {0.419} &0.400 &0.406&{0.404}&{0.408} &{0.485} &{0.481} &{0.448} &{0.452} &{0.481} &{0.456} &{0.588} &{0.517} \\
    \midrule

    \multirow{5}{*}{\rotatebox{90}{{ETTm2}}} 
    & {96} &\boldres{{0.175}}&\boldres{{0.260}}& 0.180 & 0.264 & \secondres{0.176} & \secondres{0.261} & 0.183 & 0.266 & 0.180 & 0.264 &{0.183} &{0.270}&{0.287}&{0.366}& {0.207} & {0.305} &{{0.187}} &{0.267}&{0.193}&{0.293}&{0.286} &{0.377}
    &{0.203} &{0.287} &{0.192} &{0.274} &{0.255} &{0.339} 
    \\
    & {192} &\boldres{{0.237}}&\boldres{{0.302}}&0.245 & 0.304 & \secondres{0.240} & \secondres{0.302} & 0.251 & 0.310 & {0.250} & 0.309&{0.255}&{0.314}&{0.414}&{0.492}& {0.290} & {0.364} &0.249 &{{0.309}}&{0.284}&{0.361}&{0.399} &{0.445}
    &{0.269} &{0.328} &{0.280} &{0.339} &{0.281} &{0.340} 
    \\
    & {336} &\secondres{0.303}&\secondres{0.343}& 0.316	& 0.351 & \boldres{0.299} & \boldres{0.339} & 0.319 & 0.351 & {{0.311}} & 0.348 &0.309 & 0.347&{0.597}&{0.542} & {0.377} & {0.422} &{{0.321}} &{{0.351}}&{0.382}&{0.429}&{0.637} &{0.591}
    &{0.325} &{0.366} &{0.334} &{0.361} &{0.339} &{0.372} 
    \\
    & {720} &\boldres{{0.396}}& \boldres{0.400} & 0.409 & \secondres{0.403} & \secondres{0.400} & {0.416} & 0.420 & 0.410 & 0.412 & 0.407&{{0.412}}&{0.404}&{1.730}&{1.042}& {0.558} & {0.524} &{{0.408}} &0.403&{0.558}&{0.525}&{0.960} &{0.735}
     &{0.421} &{0.415} &{0.417} &{0.413} &{0.433} &{0.432} \\
    \cmidrule(lr){2-30}
    &{Avg} &\boldres{{0.278}}&\boldres{0.326} & 0.288 & 0.330 & \secondres{0.279} & \secondres{0.329} & 0.293 & 0.334 & 0.288 & \secondres{{0.332}}&{0.290}&{0.334}&{0.757}&{0.610}& {0.358} & {0.404} &{{0.291}} &{{0.333}}&{0.354}&{0.402} &{0.954} &{0.723}&{0.305} &{0.349} &{0.306} &{0.347} &{0.327} &{0.371} \\
%    \bottomrule
\midrule
\multicolumn{2}{c|}{{$1^{st}$ Count}} & {27} & {22} & 0 & 2 & 6 & 7 & 2 & 0 & {0} & {4} & {0} & {0} & {0} & {0} & {0} & {0} & {0} & {0} & {0} & {0} & {0} & {0} & {0} & {0} & {0} & {0} & {0} & {0} \\
\bottomrule
  \end{tabular}
    \end{small}
  \end{threeparttable}
  }
\end{table}

%% file: tables/classification_full.tex
\begin{table}[tbp]
  \caption{Full results for the classification task. $\ast.$ in the model names indicates the name of $\ast$former. We report the classification accuracy (\%) as the result. The standard deviation is within 0.1\%. {Red Bold}: the best results; {underlined in blue}: the second best.}
  \label{tab:full_classification_results}
  \vskip 0.05in
  \centering
  \resizebox{\columnwidth}{!}{
  \begin{threeparttable}
  \begin{small}
  \renewcommand{\multirowsetup}{\centering}
  \setlength{\tabcolsep}{2pt}
  \begin{tabular}{c|cccccccccccccccccc}
    \toprule
    \scalebox{0.8}{Methods} & \scalebox{0.7}{DTW} & \scalebox{0.6}{XGBoost} & \scalebox{0.6}{LSTM} & \scalebox{0.6}{LSTNet} & \scalebox{0.6}{LSSL} & \scalebox{0.6}{TCN} & \scalebox{0.7}{Trans.} & \scalebox{0.7}{Re.} & \scalebox{0.7}{In.} & \scalebox{0.7}{Pyra.} & \scalebox{0.7}{Auto.} &  \scalebox{0.7}{FED.} & \scalebox{0.7}{ETS.} &  \scalebox{0.7}{iTrans.}& \scalebox{0.6}{DLinear} & \scalebox{0.6}{LightTS}&  \scalebox{0.7}{TiDE} &  \scalebox{0.7}{\textbf{MAS4TS}} \\
	\scalebox{0.8}{Datasets} & \scalebox{0.6}{\citeyearpar{Berndt1994UsingDT}} & \scalebox{0.6}{\citeyearpar{Chen2016XGBoostAS}} & \scalebox{0.6}{\citeyearpar{Hochreiter1997LongSM}} & 
	\scalebox{0.6}{\citeyearpar{2018Modeling}} & 
	\scalebox{0.6}{\citeyearpar{gu2022efficiently}} & 
	\scalebox{0.6}{\citeyearpar{Franceschi2019UnsupervisedSR}} & \scalebox{0.6}{\citeyearpar{vaswani2017attention}} & 
	\scalebox{0.6}{\citeyearpar{kitaev2020reformer}} & \scalebox{0.6}{\citeyearpar{zhou2021informer}} & \scalebox{0.6}{\citeyearpar{liu2022pyraformer}} &
	\scalebox{0.6}{\citeyearpar{wu2021autoformer}} & 
	\scalebox{0.6}{\citeyearpar{zhou2022fedformer}} & \scalebox{0.6}{\citeyearpar{woo2022etsformer}} & \scalebox{0.6}{\citeyearpar{liu2023itransformer}} & 
	\scalebox{0.6}{\citeyearpar{zeng2023dlinear}} & \scalebox{0.6}{\citeyearpar{campos2023lightts}} & \scalebox{0.6}{\citeyearpar{das2023long}}& \scalebox{0.6}{\textbf{(Ours)}} \\
    \midrule
	\scalebox{0.7}{EthanolConcentration} & \scalebox{0.8}{32.3} & {\scalebox{0.8}{43.7}} & \scalebox{0.8}{32.3} & \scalebox{0.8}{39.9} & \scalebox{0.8}{31.1}&  \scalebox{0.8}{28.9} & \scalebox{0.8}{32.7} &\scalebox{0.8}{31.9} &\scalebox{0.8}{31.6}   &\scalebox{0.8}{30.8} &\scalebox{0.8}{31.6} & \scalebox{0.8}{28.1}&\scalebox{0.8}{31.2}  & \scalebox{0.8}{28.1}& \scalebox{0.8}{32.6} &\scalebox{0.8}{29.7} & \scalebox{0.8}{27.1}& {\scalebox{0.8}{33.46}}\\
	\scalebox{0.7}{FaceDetection} & \scalebox{0.8}{52.9} & \scalebox{0.8}{63.3} & \scalebox{0.8}{57.7} & \scalebox{0.8}{65.7} & \scalebox{0.8}{66.7} & \scalebox{0.8}{52.8} & \scalebox{0.8}{67.3} & {\scalebox{0.8}{68.6}} &\scalebox{0.8}{67.0} &\scalebox{0.8}{65.7} &\scalebox{0.8}{68.4} &\scalebox{0.8}{66.0} & \scalebox{0.8}{66.3} & \scalebox{0.8}{66.3}&\scalebox{0.8}{68.0} &\scalebox{0.8}{67.5} & \scalebox{0.8}{65.3}& {\scalebox{0.8}{69.38}}  \\
	\scalebox{0.7}{Handwriting} & \scalebox{0.8}{28.6} & \scalebox{0.8}{15.8} & \scalebox{0.8}{15.2} & \scalebox{0.8}{25.8} & \scalebox{0.8}{24.6} & {\scalebox{0.8}{53.3}} & \scalebox{0.8}{32.0} & \scalebox{0.8}{27.4} &\scalebox{0.8}{32.8} &\scalebox{0.8}{29.4} &\scalebox{0.8}{36.7} &\scalebox{0.8}{28.0} &  \scalebox{0.8}{32.5} & \scalebox{0.8}{24.2}& \scalebox{0.8}{27.0} &\scalebox{0.8}{26.1} & \scalebox{0.8}{23.2}& {\scalebox{0.8}{33.64}} \\
	\scalebox{0.7}{Heartbeat} & \scalebox{0.8}{71.7}  & \scalebox{0.8}{73.2} & \scalebox{0.8}{72.2} & \scalebox{0.8}{77.1} & \scalebox{0.8}{72.7}& \scalebox{0.8}{75.6} & \scalebox{0.8}{76.1} & \scalebox{0.8}{77.1} &{\scalebox{0.8}{80.5}} &\scalebox{0.8}{75.6} &\scalebox{0.8}{74.6} &\scalebox{0.8}{73.7} &  \scalebox{0.8}{71.2} & \scalebox{0.8}{75.6}& \scalebox{0.8}{75.1} &\scalebox{0.8}{75.1} & \scalebox{0.8}{74.6}& {\scalebox{0.8}{78.04}}    \\
	\scalebox{0.7}{JapaneseVowels} & \scalebox{0.8}{94.9} & \scalebox{0.8}{86.5} & \scalebox{0.8}{79.7} & \scalebox{0.8}{98.1} & \scalebox{0.8}{98.4} & {\scalebox{0.8}{98.9}} & \scalebox{0.8}{98.7} & \scalebox{0.8}{97.8} &{\scalebox{0.8}{98.9}} &\scalebox{0.8}{98.4} &\scalebox{0.8}{96.2} &\scalebox{0.8}{98.4} & \scalebox{0.8}{95.9} & \scalebox{0.8}{96.6}& \scalebox{0.8}{96.2} &\scalebox{0.8}{96.2} & \scalebox{0.8}{95.6}& {\scalebox{0.8}{97.56}}  \\
	\scalebox{0.7}{SelfRegulationSCP1} & \scalebox{0.8}{77.7}  & \scalebox{0.8}{84.6} & \scalebox{0.8}{68.9} & \scalebox{0.8}{84.0} & \scalebox{0.8}{90.8} & \scalebox{0.8}{84.6} & {\scalebox{0.8}{92.2}} & \scalebox{0.8}{90.4} &\scalebox{0.8}{90.1} &\scalebox{0.8}{88.1} &\scalebox{0.8}{84.0} &\scalebox{0.8}{88.7} & \scalebox{0.8}{89.6} & \scalebox{0.8}{90.2}& \scalebox{0.8}{87.3} &\scalebox{0.8}{89.8} & \scalebox{0.8}{89.2}& {\scalebox{0.8}{91.81}} \\
    \scalebox{0.7}{SelfRegulationSCP2} & \scalebox{0.8}{53.9} & \scalebox{0.8}{48.9} & \scalebox{0.8}{46.6} & \scalebox{0.8}{52.8} & \scalebox{0.8}{52.2} & \scalebox{0.8}{55.6} & \scalebox{0.8}{53.9} & {\scalebox{0.8}{56.7}} &\scalebox{0.8}{53.3} &\scalebox{0.8}{53.3} &\scalebox{0.8}{50.6} &\scalebox{0.8}{54.4} & \scalebox{0.8}{55.0} & \scalebox{0.8}{54.4}& \scalebox{0.8}{50.5} &\scalebox{0.8}{51.1} & \scalebox{0.8}{53.4}& {\scalebox{0.8}{55.55}}  \\
    \scalebox{0.7}{UWaveGestureLibrary} & {\scalebox{0.8}{90.3}} & \scalebox{0.8}{75.9} & \scalebox{0.8}{41.2} & \scalebox{0.8}{87.8} & \scalebox{0.8}{85.9} & \scalebox{0.8}{88.4} & \scalebox{0.8}{85.6} & \scalebox{0.8}{85.6} &\scalebox{0.8}{85.6} &\scalebox{0.8}{83.4} &\scalebox{0.8}{85.9} &\scalebox{0.8}{85.3} & \scalebox{0.8}{85.0} & \scalebox{0.8}{85.9}& \scalebox{0.8}{82.1} &\scalebox{0.8}{80.3} & \scalebox{0.8}{84.9}& {\scalebox{0.8}{86.56}} \\
    \midrule
    \scalebox{0.8}{Average Accuracy} & \scalebox{0.8}{62.78} & \scalebox{0.8}{61.48} & \scalebox{0.8}{51.72} & \scalebox{0.8}{66.40} & \scalebox{0.8}{65.30} & \scalebox{0.8}{67.26} & \scalebox{0.8}{67.31} & {\scalebox{0.8}{66.93}} &\scalebox{0.8}{59.93} &\scalebox{0.8}{65.58} &\scalebox{0.8}{66.00} &\scalebox{0.8}{65.32} & \scalebox{0.8}{65.83} & \scalebox{0.8}{65.16}& \scalebox{0.8}{64.85} &\scalebox{0.8}{64.47} & \scalebox{0.8}{64.16}& {\scalebox{0.8}{\textbf{68.25}}} \\
	\bottomrule
  \end{tabular}
    \end{small}
  \end{threeparttable}
  }
\end{table}

%% file: tables/imputation_full.tex
\begin{table}[!ht]
\centering
\caption{Full results for the imputation task. We randomly mask 12.5\%, 25\%, 37.5\% and 50\% time points to compare the model performance under different missing degrees. $*$ in the Transformers indicates the name of $*$former.}
\label{tab:full_imputation_results}
\resizebox{\textwidth}{!}{
\begin{tabular}{cc|cc|cc|cc|cc|cc|cc|cc|cc|cc|cc|cc}
\hline
\multicolumn{2}{c|}{Models} & \multicolumn{2}{c|}{MAS4TS} & \multicolumn{2}{c|}{ETS.} & \multicolumn{2}{c|}{LightTS$^*$} & \multicolumn{2}{c|}{DLinear$^*$} & \multicolumn{2}{c|}{FED.} & \multicolumn{2}{c|}{Pyra.} & \multicolumn{2}{c|}{In.} & \multicolumn{2}{c|}{Re.} & \multicolumn{2}{c|}{LSTM} & \multicolumn{2}{c|}{TCN} & \multicolumn{2}{c}{LSSL} \\
\multicolumn{2}{c|}{} & \multicolumn{2}{c|}{(Ours)} & \multicolumn{2}{c|}{(2022)} & \multicolumn{2}{c|}{(2022)} & \multicolumn{2}{c|}{(2023)} & \multicolumn{2}{c|}{(2022)} & \multicolumn{2}{c|}{(2021a)} & \multicolumn{2}{c|}{(2021)} & \multicolumn{2}{c|}{(2020)} & \multicolumn{2}{c|}{(1997)} & \multicolumn{2}{c|}{(2019)} & \multicolumn{2}{c}{(2022)} \\
\hline
\multicolumn{2}{c|}{Mask Ratio} & MSE & MAE & MSE & MAE & MSE & MAE & MSE & MAE & MSE & MAE & MSE & MAE & MSE & MAE & MSE & MAE & MSE & MAE & MSE & MAE & MSE & MAE \\
\hline
\multirow{5}{*}{\rotatebox{90}{ETTm1}} & 12.5\% & 0.046 & 0.137 & 0.067 & 0.188 & 0.075 & 0.180 & 0.058 & 0.162 & \secondres{0.035} & \secondres{0.135} & 0.670 & 0.541 & 0.047 & 0.155 & \boldres{0.032} & \boldres{0.126} & 0.974 & 0.780 & 0.510 & 0.493 & 0.101 & 0.231 \\
& 25\% & \secondres{0.049} & \boldres{0.143} & 0.096 & 0.229 & 0.093 & 0.206 & 0.080 & 0.193 & 0.052 & 0.166 & 0.689 & 0.553 & 0.063 & 0.180 & \boldres{0.042} & \secondres{0.146} & 1.032 & 0.807 & 0.518 & 0.500 & 0.106 & 0.235 \\
& 37.5\% & \secondres{0.054} & \boldres{0.150} & 0.133 & 0.271 & 0.113 & 0.231 & 0.103 & 0.219 & 0.069 & 0.191 & 0.737 & 0.581 & 0.079 & 0.200 & \boldres{0.052} & \secondres{0.158} & 0.999 & 0.792 & 0.516 & 0.499 & 0.116 & 0.246 \\
& 50\% & \secondres{0.061} & \boldres{0.159} & 0.186 & 0.323 & 0.134 & 0.255 & 0.132 & 0.248 & 0.089 & 0.218 & 0.770 & 0.605 & 0.093 & 0.218 & \boldres{0.063} & \secondres{0.173} & 0.952 & 0.763 & 0.519 & 0.496 & 0.129 & 0.260 \\
& Avg & \secondres{0.052} & \boldres{0.147} & 0.120 & 0.253 & 0.104 & 0.218 & 0.093 & 0.206 & 0.062 & 0.177 & 0.717 & 0.570 & 0.071 & 0.188 & \boldres{0.050} & \secondres{0.154} & 0.989 & 0.786 & 0.516 & 0.497 & 0.113 & 0.254 \\
\hline
\multirow{5}{*}{\rotatebox{90}{ETTm2}} & 12.5\% & \boldres{0.026} & \boldres{0.095} & 0.108 & 0.239 & \secondres{0.034} & \secondres{0.127} & 0.062 & 0.166 & 0.056 & 0.159 & 0.394 & 0.470 & 0.133 & 0.270 & 0.108 & 0.228 & 1.013 & 0.805 & 0.307 & 0.441 & 0.150 & 0.298 \\
& 25\% & \boldres{0.029} & \boldres{0.100} & 0.164 & 0.294 & \secondres{0.042} & \secondres{0.143} & 0.085 & 0.196 & 0.080 & 0.195 & 0.421 & 0.482 & 0.135 & 0.272 & 0.136 & 0.262 & 1.039 & 0.814 & 0.263 & 0.402 & 0.159 & 0.306 \\
& 37.5\% & \boldres{0.031} & \boldres{0.105} & 0.237 & 0.356 & \secondres{0.051} & \secondres{0.159} & 0.106 & 0.222 & 0.110 & 0.231 & 0.478 & 0.521 & 0.155 & 0.293 & 0.175 & 0.300 & 0.917 & 0.744 & 0.250 & 0.396 & 0.180 & 0.321 \\
& 50\% & \boldres{0.034} & \boldres{0.111} & 0.323 & 0.421 & \secondres{0.059} & \secondres{0.174} & 0.131 & 0.247 & 0.156 & 0.276 & 0.568 & 0.560 & 0.200 & 0.333 & 0.211 & 0.329 & 1.140 & 0.835 & 0.246 & 0.389 & 0.210 & 0.353 \\
& Avg & \boldres{0.030} & \boldres{0.103} & 0.208 & 0.327 & \secondres{0.046} & \secondres{0.151} & 0.096 & 0.208 & 0.101 & 0.215 & 0.465 & 0.508 & 0.156 & 0.292 & 0.157 & 0.280 & 1.027 & 0.800 & 0.266 & 0.407 & 0.175 & 0.324 \\
\hline
\multirow{5}{*}{\rotatebox{90}{ETTh1}} & 12.5\% & 0.104 & 0.211 & 0.126 & 0.263 & 0.240 & 0.345 & 0.151 & 0.267 & \boldres{0.070} & \boldres{0.190} & 0.857 & 0.609 & 0.114 & 0.234 & \secondres{0.074} & \secondres{0.194} & 1.265 & 0.896 & 0.599 & 0.554 & 0.422 & 0.461 \\
& 25\% & 0.119 & \boldres{0.227} & 0.169 & 0.304 & 0.265 & 0.364 & 0.180 & 0.292 & \secondres{0.106} & 0.236 & 0.829 & 0.672 & 0.140 & 0.262 & \boldres{0.102} & \secondres{0.227} & 1.262 & 0.883 & 0.610 & 0.567 & 0.412 & 0.456 \\
& 37.5\% & 0.137 & \boldres{0.243} & 0.220 & 0.347 & 0.296 & 0.382 & 0.215 & 0.318 & \boldres{0.124} & \secondres{0.258} & 0.830 & 0.675 & 0.174 & 0.293 & \secondres{0.135} & 0.261 & 1.200 & 0.867 & 0.628 & 0.577 & 0.421 & 0.461 \\
& 50\% & \boldres{0.158} & \boldres{0.261} & 0.293 & 0.402 & 0.334 & 0.404 & 0.257 & 0.347 & \secondres{0.165} & \secondres{0.299} & 0.854 & 0.691 & 0.215 & 0.325 & 0.179 & \secondres{0.298} & 1.174 & 0.849 & 0.648 & 0.587 & 0.443 & 0.473 \\
& Avg & 0.129 & \boldres{0.235} & 0.202 & 0.329 & 0.284 & 0.373 & 0.201 & 0.306 & \boldres{0.117} & 0.246 & 0.842 & 0.682 & 0.161 & 0.279 & \secondres{0.122} & \secondres{0.245} & 1.225 & 0.873 & 0.621 & 0.571 & 0.424 & 0.481 \\
\hline
\multirow{5}{*}{\rotatebox{90}{ETTh2}} & 12.5\% & \boldres{0.062} & \boldres{0.161} & 0.187 & 0.319 & 0.101 & 0.231 & 0.100 & 0.216 & \secondres{0.095} & \secondres{0.212} & 0.976 & 0.754 & 0.305 & 0.431 & 0.163 & 0.289 & 2.060 & 1.120 & 0.410 & 0.494 & 0.521 & 0.555 \\
& 25\% & \boldres{0.063} & \boldres{0.160} & 0.279 & 0.390 & \secondres{0.115} & \secondres{0.246} & 0.127 & 0.247 & 0.137 & 0.258 & 1.037 & 0.774 & 0.322 & 0.444 & 0.206 & 0.331 & 2.007 & 1.105 & 0.419 & 0.490 & 0.487 & 0.535 \\
& 37.5\% & \boldres{0.068} & \boldres{0.168} & 0.400 & 0.465 & \secondres{0.126} & \secondres{0.257} & 0.158 & 0.276 & 0.187 & 0.304 & 1.107 & 0.800 & 0.353 & 0.462 & 0.252 & 0.370 & 2.033 & 1.111 & 0.429 & 0.498 & 0.487 & 0.529 \\
& 50\% & \boldres{0.074} & \boldres{0.176} & 0.602 & 0.572 & \secondres{0.136} & \secondres{0.268} & 0.183 & 0.299 & 0.232 & 0.341 & 1.193 & 0.838 & 0.369 & 0.472 & 0.316 & 0.419 & 2.054 & 1.119 & 0.467 & 0.529 & 0.484 & 0.523 \\
& Avg & \boldres{0.067} & \boldres{0.166} & 0.367 & 0.436 & \secondres{0.119} & \secondres{0.250} & 0.142 & 0.259 & 0.163 & 0.279 & 1.079 & 0.792 & 0.337 & 0.452 & 0.234 & 0.352 & 2.039 & 1.114 & 0.431 & 0.503 & 0.495 & 0.475 \\
\hline
\multirow{5}{*}{\rotatebox{90}{Electricity}} & 12.5\% & \boldres{0.070} & \boldres{0.188} & 0.196 & 0.321 & 0.102 & 0.229 & \secondres{0.092} & \secondres{0.214} & 0.107 & 0.237 & 0.297 & 0.383 & 0.218 & 0.326 & 0.190 & 0.308 & 0.277 & 0.366 & 0.621 & 0.620 & 0.217 & 0.341 \\
& 25\% & \boldres{0.079} & \boldres{0.199} & 0.207 & 0.332 & 0.121 & 0.252 & \secondres{0.118} & \secondres{0.247} & 0.120 & 0.251 & 0.294 & 0.380 & 0.219 & 0.326 & 0.197 & 0.312 & 0.281 & 0.369 & 0.559 & 0.585 & 0.219 & 0.341 \\
& 37.5\% & \boldres{0.090} & \boldres{0.201} & 0.219 & 0.344 & 0.141 & 0.273 & 0.144 & 0.276 & \secondres{0.136} & \secondres{0.266} & 0.296 & 0.381 & 0.222 & 0.328 & 0.203 & 0.315 & 0.275 & 0.364 & 0.567 & 0.588 & 0.223 & 0.343 \\
& 50\% & \boldres{0.107} & \boldres{0.221} & 0.235 & 0.357 & 0.160 & 0.293 & 0.175 & 0.305 & \secondres{0.158} & \secondres{0.284} & 0.299 & 0.383 & 0.228 & 0.331 & 0.210 & 0.319 & 0.273 & 0.361 & 0.581 & 0.597 & 0.229 & 0.347 \\
& Avg & \boldres{0.086} & \boldres{0.202} & 0.214 & 0.339 & 0.131 & 0.262 & 0.132 & 0.260 & \secondres{0.130} & \secondres{0.259} & 0.297 & 0.382 & 0.222 & 0.328 & 0.200 & 0.313 & 0.277 & 0.365 & 0.582 & 0.597 & 0.222 & 0.293 \\
\hline
\multirow{5}{*}{\rotatebox{90}{Weather}} & 12.5\% & \secondres{0.035} & 0.090 & 0.057 & 0.141 & 0.047 & 0.101 & 0.039 & \secondres{0.084} & 0.041 & 0.107 & 0.140 & 0.220 & 0.037 & 0.093 & \boldres{0.031} & \boldres{0.076} & 0.296 & 0.379 & 0.176 & 0.287 & 0.036 & 0.095 \\
& 25\% & \boldres{0.035} & \boldres{0.062} & 0.065 & 0.155 & 0.052 & 0.111 & 0.048 & 0.103 & 0.064 & 0.163 & 0.147 & 0.229 & 0.042 & 0.100 & \secondres{0.035} & \secondres{0.082} & 0.327 & 0.409 & 0.187 & 0.293 & 0.042 & 0.104 \\
& 37.5\% & \boldres{0.037} & \boldres{0.065} & 0.081 & 0.180 & 0.058 & 0.121 & 0.057 & 0.117 & 0.107 & 0.229 & 0.156 & 0.240 & 0.049 & 0.111 & \secondres{0.040} & \secondres{0.091} & 0.406 & 0.463 & 0.172 & 0.281 & 0.047 & 0.112 \\
& 50\% & \boldres{0.036} & \boldres{0.063} & 0.102 & 0.207 & 0.065 & 0.133 & 0.066 & 0.134 & 0.183 & 0.312 & 0.164 & 0.249 & 0.053 & 0.114 & \secondres{0.046} & \secondres{0.099} & 0.431 & 0.483 & 0.195 & 0.303 & 0.054 & 0.123 \\
& Avg & \boldres{0.036} & \boldres{0.070} & 0.076 & 0.171 & 0.055 & 0.117 & 0.052 & 0.110 & 0.099 & 0.203 & 0.152 & 0.235 & 0.045 & 0.104 & \secondres{0.038} & \secondres{0.087} & 0.365 & 0.434 & 0.183 & 0.291 & 0.045 & 0.108 \\
\hline
\multirow{5}{*}{\rotatebox{90}{Exchange}} & 12.5\% & \boldres{0.004} & \boldres{0.038} & 0.043 & 0.151 & 0.040 & 0.137 & 0.023 & 0.106 & 0.112 & 0.224 & 0.235 & 0.407 & 0.345 & 0.483 & 0.280 & 0.411 & \secondres{0.017} & \secondres{0.090} & 0.025 & 0.112 & 0.019 & 0.098 \\
& 25\% & \boldres{0.004} & \boldres{0.039} & 0.104 & 0.240 & 0.043 & 0.143 & 0.037 & 0.134 & 0.184 & 0.288 & 0.412 & 0.551 & 0.443 & 0.561 & 0.253 & 0.383 & \secondres{0.030} & \secondres{0.122} & 0.037 & 0.136 & 0.034 & 0.130 \\
& 37.5\% & \boldres{0.005} & \boldres{0.042} & 0.149 & 0.296 & 0.045 & 0.148 & 0.053 & 0.159 & 0.281 & 0.355 & 0.447 & 0.579 & 0.393 & 0.522 & 0.316 & 0.472 & \secondres{0.045} & \secondres{0.150} & 0.050 & 0.159 & 0.046 & 0.153 \\
& 50\% & \boldres{0.005} & \boldres{0.046} & 0.215 & 0.363 & \secondres{0.039} & 0.140 & 0.071 & 0.185 & 0.387 & 0.431 & 0.499 & 0.593 & 0.575 & 0.649 & 0.358 & 0.489 & 0.036 & \secondres{0.130} & 0.061 & 0.176 & 0.062 & 0.176 \\
& Avg & \boldres{0.005} & \boldres{0.041} & 0.128 & 0.262 & 0.042 & 0.142 & 0.046 & 0.146 & 0.241 & 0.324 & 0.398 & 0.532 & 0.439 & 0.554 & 0.302 & 0.439 & \secondres{0.032} & \secondres{0.123} & 0.043 & 0.146 & 0.040 & 0.139 \\
\hline
\multicolumn{2}{c|}{1st Count} & 25 & 32 & 0 & 0 & 0 & 0 & 0 & 0 & 3 & 1 & 0 & 0 & 0 & 0 & 7 & 2 & 0 & 0 & 0 & 0 & 0 & 0 \\
\hline
\end{tabular}
}
\end{table}

%% file: tables/anomaly_full.tex
\begin{table*}[h]
\centering
\small
\caption{Full results for the anomaly detection task. The P, R and F1 represent the precision, recall and F1-score (\%) respectively. F1-score is the harmonic mean of precision and recall. A higher value of P, R and F1 indicates a better performance.}
\label{tab:full_anomaly_detection_results}
\resizebox{\textwidth}{!}{
\begin{tabular}{c|ccc|ccc|ccc|ccc|ccc|c}
\toprule
{Models} &
\multicolumn{3}{c|}{\textbf{SMD}} &
\multicolumn{3}{c|}{\textbf{MSL}} &
\multicolumn{3}{c|}{\textbf{SMAP}} &
\multicolumn{3}{c|}{\textbf{SWaT}} &
\multicolumn{3}{c|}{\textbf{PSM}} &
\multirow{2}{*}{\textbf{Avg F1}} \\

\cmidrule(lr){2-4}
\cmidrule(lr){5-7}
\cmidrule(lr){8-10}
\cmidrule(lr){11-13}
\cmidrule(lr){14-16}
Metric & P & R & F1 & P & R & F1 & P & R & F1 & P & R & F1 & P & R & F1 & \\
\midrule
\scalebox{0.9}{LSTM (1997)} \scalebox{0.8}{\citeyear{hochreiter1997long}} & 78.52 & 65.47 & 71.41 & 78.04 & 86.22 & 81.93 & 91.06 & 57.49 & 70.48 & 78.06 & 91.72 & 84.34 & 69.24 & 99.53 & 81.67 & 77.97 \\
\scalebox{0.9}{Transformer} \scalebox{0.8}{\citeyear{vaswani2017attention}} & 83.58 & 76.13 & 79.56 & 71.57 & 87.37 & 78.68 & 89.37 & 57.12 & 69.70 & 68.84 & 96.53 & 80.37 & 62.75 & 96.56 & 76.07 & 76.88 \\
\scalebox{0.9}{LogTrans} \scalebox{0.8}{\citeyear{li2019enhancing}} & 83.46 & 70.13 & 76.21 & 73.05 & 87.37 & 79.57 & 89.15 & 57.59 & 69.97 & 68.67 & 97.32 & 80.52 & 63.06 & 98.00 & 76.74 & 76.60 \\
\scalebox{0.9}{TCN} \scalebox{0.8}{\citeyear{Franceschi2019UnsupervisedSR}}& 84.06 & 79.07 & 81.49 & 75.11 & 82.44 & 78.60 & 86.90 & 59.23 & 70.45 & 76.59 & 95.71 & 85.09 & 54.59 & 99.77 & 70.57 & 77.24 \\
\scalebox{0.9}{Reformer} \scalebox{0.8}{\citeyear{kitaev2020reformer}} & 82.58 & 69.24 & 75.32 & 85.51 & 83.31 & 84.40 & 90.91 & 57.44 & 70.40 & 72.50 & 96.53 & 82.80 & 59.93 & 95.38 & 73.61 & 77.31 \\
\scalebox{0.9}{Informer} \scalebox{0.8}{\citeyear{zhou2021informer}} & 86.60 & 77.23 & 81.65 & 81.77 & 86.48 & 84.06 & 90.11 & 57.13 & 69.92 & 70.29 & 96.75 & 81.43 & 64.27 & 96.33 & 77.10 & 78.83 \\
\scalebox{0.9}{Autoformer} \scalebox{0.8}{\citeyear{wu2021autoformer}} 88.06 & 82.35 & 85.11 & 77.27 & 80.92 & 79.05 & 90.40 & 58.62 & 71.12 & 89.85 & 95.81 & 92.74 & 99.08 & 88.15 & 93.29 & 84.26 \\
\scalebox{0.9}{Pyraformer} \scalebox{0.8}{\citeyear{liu2022pyraformer}} & 85.61 & 80.61 & 83.04 & 83.81 & 85.93 & 84.86 & 92.54 & 57.71 & 71.09 & 87.92 & 96.00 & 91.78 & 71.67 & 96.02 & 82.08 & 82.57 \\
\scalebox{0.9}{Anomaly Trans.} \scalebox{0.8}{\citeyear{xu2021anomaly}} & 88.91 & 82.23 & 85.49 & 79.61 & 87.37 & 83.31 & 91.85 & 58.11 & 71.18 & 72.51 & 97.32 & 83.10 & 68.35 & 94.72 & 79.40 & 80.50 \\
\scalebox{0.9}{Stationary} \scalebox{0.8}{\citeyear{liu2022non}} & 88.33 & 81.21 & 84.62 & 68.55 & 89.14 & 77.50 & 89.37 & 59.02 & 71.09 & 68.03 & 96.75 & 79.88 & 97.82 & 96.76 & 97.29 & 82.08 \\
\scalebox{0.9}{LSSL} \scalebox{0.8}{\citeyear{gu2022efficiently}} & 78.51 & 65.32 & 71.31 & 77.55 & 88.18 & 82.53 & 89.43 & 53.43 & 66.90 & 79.05 & 93.72 & 85.76 & 66.02 & 92.93 & 77.20 & 76.74 \\
\scalebox{0.9}{ETSformer} \scalebox{0.8}{\citeyear{woo2022etsformer}} & 87.44 & 79.23 & 83.13 & 85.13 & 84.93 & 85.03 & 92.25 & 55.75 & 69.50 & 90.02 & 80.36 & 84.91 & 99.31 & 85.28 & 91.76 & 82.87 \\
\scalebox{0.9}{DLinear} \scalebox{0.8}{\citeyear{zeng2023dlinear}} & 83.62 & 71.52 & 77.10 & 84.34 & 85.42 & 84.88 & 92.32 & 55.41 & 69.26 & 80.91 & 95.30 & 87.52 & 98.28 & 89.26 & 93.55 & 82.46 \\
\scalebox{0.9}{TiDE} \scalebox{0.8}{\citeyear{das2023long}} & 76.00 & 63.00 & 68.91 & 84.00 & 60.00 & 70.18 & 88.00 & 50.00 & 64.00 & 98.00 & 63.00 & 76.73 & 93.00 & 92.00 & 92.50 & 74.46 \\
\scalebox{0.9}{PatchTST} \scalebox{0.8}{\citeyear{nie2023patchtst}} & 87.26 & 82.14 & 84.62 & 88.34 & 70.96 & 78.70 & 90.64 & 55.46 & 68.82 & 91.10 & 80.94 & 85.72 & 98.84 & 93.47 & 96.08 & 82.79 \\
\scalebox{0.9}{iTransformer} \scalebox{0.8}{\citeyear{liu2023itransformer}} & 78.45 & 65.10 & 71.15 & 86.15 & 62.65 & 72.54 & 90.67 & 52.96 & 66.87 & 99.96 & 65.55 & 79.18 & 95.65 & 94.69 & 95.17 & 76.98 \\
\midrule
\scalebox{0.9}{\textbf{MAS4TS (Ours)}} & 86.76 & 83.51 & 85.10 & 90.65 & 74.96 & 82.06 & 89.87 & 55.44 & 68.58 & 91.61 & 94.74 & 93.14 & 91.87 & 98.44 & 95.04 & \textbf{84.78} \\
\bottomrule
\end{tabular}
}
\end{table*}

%% file: appendix/tool_library.tex
\newpage
\section{Tool Library and Adaptive Selection}
\label{appx:tool_library}

This appendix provides implementation details for the tool library $\mathcal{T}$ introduced in Section~\ref{sec:execution_optimization}. We describe the unified invocation interface, fundamental building blocks, task-specific tool architectures, and the adaptive routing mechanism. In particular, we clarify (i) the \emph{candidate chain space} considered by the router, (ii) the \emph{search/selection procedure} at inference, and (iii) the \emph{training objective} used for learning the routing policy, so that the tool-driven execution can be reproduced and audited.

%==============================================================================
\subsection{Unified Tool Interface}
\label{appx:unified_interface}

All tools in MAS4TS expose a standardized invocation interface compatible with the routing policy $\pi_{\Theta}$ defined in the main text. 
%The tool chain is \emph{not} fixed but dynamically composed based on routing confidence:
% \begin{lstlisting}[language=Python, basicstyle=\small\ttfamily]
% # Adaptive tool invocation
% response = tool_router.call(
%     task="forecasting",
%     context={
%         "data": x_enc,           # Input sequence
%         "priors": priors,        # Structured priors P
%         "memory": shared_memory  # Shared memory M
%     }
%     # Returns: (selected_chain, confidence)
%     # e.g., ["RevIN", "PatchEmbed", "Transformer", "Head"], 0.82
% )
% \end{lstlisting}
The router evaluates candidate tool chains and returns the highest-confidence selection. For instance, given strong periodicity signals, it may select \texttt{[TimesBlock, FFT]} over \texttt{[PatchEmbed, Transformer]}; given sparse anchors, it may prepend \texttt{[ODE\_Reconstruct]} to the chain.

\paragraph{Semantic-Adaptive Routing.}
The router maintains a learnable policy network $g_{\psi}$ that scores each candidate tool chain based on three inputs: the task type $\tau$, structured priors $\mathcal{P}$ extracted by reasoning agents (e.g., detected trends, anchor density, periodicity strength), and a compressed summary of the shared memory state $\mathbf{M}$. %At inference, the router computes a confidence score for each registered chain and selects the one with maximum confidence. When the top confidence falls below a threshold (empirically 0.6), the system triggers a fallback to a default chain or combines multiple high-scoring chains via ensemble weighting, ensuring robust selection under ambiguous inputs.

\textbf{Candidate chain space.}
For each task $\tau$, we maintain a registry $\mathcal{C}_\tau=\{\mathsf{Chain}_1,\ldots,\mathsf{Chain}_{|\mathcal{C}_\tau|}\}$ of candidate tool chains.
Each chain is a short sequence of tool \emph{modules} selected from the task-specific tool set (Sections~\ref{appx:forecasting_tools}--\ref{appx:anomaly_tools}) and a small set of shared preprocess/postprocess operators (e.g., normalization and masking).
To keep routing efficient and avoid combinatorial explosion, we constrain the chain length to a small integer range:
\begin{equation}
K \in [K_{\min}, K_{\max}],
\end{equation}
with $K_{\min}=1$ and $K_{\max}=3$ in our experiments.\footnote{The bounds are chosen to cover common patterns such as ``normalize $\rightarrow$ backbone $\rightarrow$ head'' while allowing one or two optional modules (e.g., decomposition, ODE reconstruction).}
Within a task, we \emph{do not} enumerate all possible compositions; instead, $\mathcal{C}_\tau$ consists of a curated set of valid, numerically stable chains that respect module I/O schemas (Section~\ref{appx:tool_chaining}).

\textbf{Search and selection at inference.}
Given an input context $(\mathbf{X},\mathcal{P},\mathbf{M})$ and a task $\tau$, the router computes a score vector over candidate chains:
\begin{equation}
\mathbf{s} = g_{\psi}\big(\tau,\mathcal{P},\text{Pool}(\mathbf{M})\big)\in\mathbb{R}^{|\mathcal{C}_\tau|},
\end{equation}
followed by a softmax to obtain $\pi_{\Theta}(\mathsf{Chain}_i \mid \tau,\mathcal{P},\mathbf{M})$ (consistent with Eq.~(9) in the main text).
We then select either (i) the top-1 chain (greedy) or (ii) a small top-$k$ set for ensembling:
\begin{equation}
\mathcal{I}_k = \text{TopK}(\mathbf{s},k), \qquad k\in\{1,2,3\}.
\end{equation}
Unless otherwise stated, we use greedy top-1 selection when the top confidence is above the threshold (0.6) and switch to top-$k$ ensembling when it is below (Appendix~\ref{appx:tool_chaining}).

\subsection{Fundamental Building Blocks}
\label{appx:component_taxonomy}

Before describing task-specific tools, we introduce the fundamental components that compose our tool library. These modules are shared across tasks and can be combined in different configurations. Here are some of the key ingredients, not all of them. And we will updated the library in the future.

\subsubsection{Patch Embedding}
\label{appx:embedding}
The patch embedding module converts continuous time series into discrete tokens for transformer-based processing:
\begin{equation}
    \mathbf{P} = \text{Linear}(\text{Unfold}(\text{Pad}(\mathbf{X}))) + \mathbf{E}_{pos},
\end{equation}
where $\text{Unfold}(\cdot)$ extracts overlapping patches with length $p$ and stride $s$, and $\mathbf{E}_{pos}$ denotes learnable positional embeddings. Default hyperparameters are \texttt{patch\_len}$=16$, \texttt{stride}$=8$, and $d=64$. This module captures local semantic patterns that point-wise tokenization cannot represent, significantly improving temporal pattern learning.

\subsubsection{Multi-Head Self-Attention}
\label{appx:mhsa}
Our attention mechanism follows the standard transformer formulation:
\begin{equation}
    \text{Attention}(\mathbf{Q}, \mathbf{K}, \mathbf{V}) = \text{softmax}\left(\frac{\mathbf{Q}\mathbf{K}^\top}{\sqrt{d_k}}\right)\mathbf{V},
\end{equation}
with multi-head extension that projects queries, keys, and values into $h$ subspaces, enabling the model to attend to information from different representation spaces.

\subsubsection{TimesBlock}
\label{appx:timesblock}
For tasks requiring periodic pattern capture (e.g., classification), we employ a specialized block that: (i) detects top-$k$ periods using FFT amplitude analysis, (ii) reshapes the 1D series to 2D representation based on detected periods, (iii) applies 2D Inception-style convolution with multiple kernel sizes, and (iv) aggregates results with learned period weights:
\begin{equation}
    \text{Period}(\mathbf{X}) = \text{FFT}^{-1}\left(\text{TopK}\left(|\text{FFT}(\mathbf{X})|, k\right)\right).
\end{equation}

\subsubsection{Normalization Strategies}
\label{appx:normalization}
We provide multiple normalization strategies optimized for non-stationary time series:
\begin{itemize}[leftmargin=*, itemsep=1pt]
    \item \textbf{RevIN} (Reversible Instance Normalization): Normalizes each instance independently and maintains statistics for denormalization, critical for preserving the original scale in forecasting.
    \item \textbf{Channel-Independent Normalization}: Normalizes each feature channel separately to avoid spurious cross-channel correlations.
    \item \textbf{Robust Normalization}: Uses percentiles instead of mean/std, providing resistance to outliers in anomaly detection.
\end{itemize}

\subsubsection{Projection Heads}
\label{appx:projection_heads}
Task-specific projection heads map latent representations to output space:
\begin{itemize}[leftmargin=*, itemsep=1pt]
    \item \textbf{FlattenHead}: Flattens patch features and projects to output dimension, supporting both channel-independent and shared modes.
    \item \textbf{MLPHead}: Two-layer MLP with configurable activation (ReLU, GELU).
    \item \textbf{ResidualHead}: Includes skip connections for stable gradient flow.
\end{itemize}

%==============================================================================
\subsection{Forecasting Tools}
\label{appx:forecasting_tools}

Table~\ref{tab:forecasting_tools} summarizes available forecasting tools. The adaptive router selects among these based on data characteristics inferred from priors $\mathcal{P}$.

\begin{table}[h]
\centering
\caption{Forecasting Tools Overview}
\label{tab:forecasting_tools}
\resizebox{\textwidth}{!}{
\begin{tabular}{llll}
\toprule
\textbf{Tool} & \textbf{Architecture} & \textbf{Key Components} & \textbf{Selection Condition} \\
\midrule
\texttt{ForecastingTool} & Patch-Transformer & PatchEmbed + Encoder + FlattenHead & Default; general-purpose \\
\texttt{DecompositionTool} & Trend-Seasonal & MovingAverage + Projection & Strong trend detected \\
\texttt{NBeatsModel} & Residual Stacks & Basis Expansion + Doubly Residual & Interpretability required \\
\bottomrule
\end{tabular}
}
\end{table}

\subsubsection{Patch-Transformer Forecasting Tool}
This is the primary forecasting tool, combining patch embedding with transformer encoding:

\begin{algorithm}[H]
\caption{Patch-Transformer Forecasting}
\label{alg:patch_forecast}
\begin{algorithmic}[1]
\REQUIRE Input $\mathbf{X} \in \mathbb{R}^{B \times L \times D}$, prediction horizon $H$
\STATE $\mathbf{X}_{norm}, \mu, \sigma \leftarrow \text{RevIN}(\mathbf{X})$ \COMMENT{Instance normalization}
\STATE $\mathbf{P} \leftarrow \text{PatchEmbed}(\mathbf{X}_{norm})$ \COMMENT{$\mathbb{R}^{B \cdot D \times N_p \times d}$}
\STATE $\mathbf{H} \leftarrow \text{TransformerEncoder}(\mathbf{P})$ \COMMENT{Multi-layer attention + FFN}
\STATE $\mathbf{Y}_{norm} \leftarrow \text{FlattenHead}(\mathbf{H})$ \COMMENT{Channel-independent projection}
\STATE $\mathbf{Y} \leftarrow \text{RevIN}^{-1}(\mathbf{Y}_{norm}, \mu, \sigma)$ \COMMENT{Denormalize}
\RETURN $\mathbf{Y} \in \mathbb{R}^{B \times H \times D}$
\end{algorithmic}
\end{algorithm}

\textbf{Channel Independence}: Each feature channel is processed independently, which improves robustness when channels have heterogeneous statistical properties and enables efficient parallel computation.

\subsubsection{Decomposition Tool}
When strong trend is detected (slope $> 0.01$), the router prepends decomposition:
\begin{equation}
    \mathbf{X}_{\text{trend}} = \text{AvgPool}_{k}(\mathbf{X}), \quad \mathbf{X}_{\text{seasonal}} = \mathbf{X} - \mathbf{X}_{\text{trend}},
\end{equation}
where $k$ is the moving average kernel size. The seasonal component is then processed by the main forecasting tool.

%==============================================================================
\subsection{Classification Tools}
\label{appx:classification_tools}

\begin{table}[h]
\centering
\caption{Classification Tools Overview}
\label{tab:classification_tools}
\resizebox{\textwidth}{!}{
\begin{tabular}{llll}
\toprule
\textbf{Tool} & \textbf{Architecture} & \textbf{Key Components} & \textbf{Selection Condition} \\
\midrule
\texttt{ClassificationTool} & TimesBlock-based & 1D Conv + FFT Period + 2D Conv & Default; periodic patterns \\
\texttt{TemporalConvNet} & Dilated Convolutions & Multi-scale 1D Conv + GlobalPool & Short sequences ($L \leq 32$) \\
\texttt{InceptionTime} & Multi-branch CNN & Inception Modules + Residual & Multi-scale features needed \\
\bottomrule
\end{tabular}
}
\end{table}

\subsubsection{TimesBlock Classification Tool}
The primary classification tool leverages FFT-based period discovery:
\begin{enumerate}[leftmargin=*, itemsep=1pt]
    \item \textbf{Data Embedding}: 1D convolution (not linear) to capture local patterns, combined with positional encoding.
    \item \textbf{TimesBlock Stack}: Multiple blocks that detect periods via FFT, reshape to 2D, apply Inception-style 2D convolution, then reshape back.
    \item \textbf{Padding Mask}: Critical for variable-length sequences---zeros out padding before flattening.
    \item \textbf{Full Sequence Projection}: Flattens $[\text{seq\_len} \times d]$ and projects to class logits.
\end{enumerate}

\subsubsection{Temporal Convolutional Network Tool}
For short sequences, the router selects TCN with dilated causal convolutions:
\begin{equation}
    (\mathbf{X} *_r f)(t) = \sum_{i=0}^{k-1} f(i) \cdot \mathbf{X}_{t - r \cdot i},
\end{equation}
where $r$ is the dilation factor that increases exponentially with depth, enabling efficient long-range dependency capture with $O(\log L)$ layers.

%==============================================================================
\subsection{Imputation Tools}
\label{appx:imputation_tools}

\begin{table}[h]
\centering
\caption{Imputation Tools Overview}
\label{tab:imputation_tools}
\resizebox{\textwidth}{!}{
\begin{tabular}{llll}
\toprule
\textbf{Tool} & \textbf{Architecture} & \textbf{Key Components} & \textbf{Selection Condition} \\
\midrule
\texttt{ImputationTool} & Patch-Transformer & PatchEmbed + Encoder + Reconstruction & Default; general-purpose \\
\texttt{SAITS} & Self-Attention & Diagonal Masked Attention + ORT & Sparse missing patterns \\
\texttt{BRITS} & Bidirectional RNN & Forward + Backward GRU + Decay & Sequential missing blocks \\
\bottomrule
\end{tabular}
}
\end{table}

\subsubsection{Patch-Transformer Imputation Tool}
Adapts the patch-based architecture for reconstruction:

\begin{algorithm}[H]
\caption{Patch-Transformer Imputation}
\label{alg:patch_impute}
\begin{algorithmic}[1]
\REQUIRE Input $\mathbf{X}$ with missing values, binary mask $\mathbf{M}$
\STATE $\mathbf{X}_{masked} \leftarrow \mathbf{X} \odot (1 - \mathbf{M})$ \COMMENT{Zero out missing positions}
\STATE $\mathbf{X}_{norm}, \mu, \sigma \leftarrow \text{Normalize}(\mathbf{X}_{masked})$
\STATE $\mathbf{P} \leftarrow \text{PatchEmbed}(\mathbf{X}_{norm})$
\STATE $\mathbf{H} \leftarrow \text{TransformerEncoder}(\mathbf{P})$
\STATE $\hat{\mathbf{X}}_{norm} \leftarrow \text{ReconstructionHead}(\mathbf{H})$ \COMMENT{Same length as input}
\STATE $\hat{\mathbf{X}} \leftarrow \text{Denormalize}(\hat{\mathbf{X}}_{norm}, \mu, \sigma)$
\STATE $\mathbf{Y} \leftarrow \mathbf{X} \odot (1 - \mathbf{M}) + \hat{\mathbf{X}} \odot \mathbf{M}$ \COMMENT{Preserve known values}
\RETURN $\mathbf{Y}$
\end{algorithmic}
\end{algorithm}

%==============================================================================
\subsection{Anomaly Detection Tools}
\label{appx:anomaly_tools}

\begin{table}[h]
\centering
\caption{Anomaly Detection Tools Overview}
\label{tab:anomaly_tools}
\resizebox{\textwidth}{!}{
\begin{tabular}{llll}
\toprule
\textbf{Tool} & \textbf{Architecture} & \textbf{Key Components} & \textbf{Selection Condition} \\
\midrule
\texttt{AnomalyDetectionTool} & Multi-scale Patch & Multi-scale PatchEmbed + Reconstruction & Default; general anomalies \\
\texttt{VAEAnomalyDetector} & VAE & Encoder + Reparameterization + Decoder & Point anomaly focus \\
\bottomrule
\end{tabular}
}
\end{table}

\subsubsection{Multi-Scale Anomaly Detection Tool}
The primary anomaly detection tool uses multi-scale reconstruction:
\begin{enumerate}[leftmargin=*, itemsep=1pt]
    \item \textbf{Instance Normalization}: Uses learnable affine parameters to preserve relative anomaly magnitude (unlike RevIN which removes it).
    \item \textbf{Multi-Scale Processing}: Fine scale ($p=4$) for point anomalies; coarse scale ($p=16$) for segment anomalies.
    \item \textbf{Multi-Scale Fusion}: Concatenates reconstructions from all scales and fuses through MLP.
\end{enumerate}

The anomaly score combines reconstruction error with attention-based weighting:
\begin{equation}
    s_t = \|\mathbf{x}_t - \hat{\mathbf{x}}_t\|_2^2 \cdot \alpha_t,
\end{equation}
where $\alpha_t$ is the attention weight highlighting suspicious regions.

\subsubsection{VAE Anomaly Detector}
Probabilistic detection through reconstruction likelihood:
\begin{align}
    \mathbf{z} &\sim q_\phi(\mathbf{z}|\mathbf{x}) = \mathcal{N}(\mu_\phi(\mathbf{x}), \sigma_\phi^2(\mathbf{x})), \\
    s &= \|\mathbf{x} - \hat{\mathbf{x}}\|_2^2 + \beta \cdot D_{KL}(q_\phi \| p),
\end{align}
where the anomaly score combines reconstruction error and KL divergence.

%==============================================================================
\subsection{Tool Chaining}
\label{appx:tool_chaining}

As described in Section~\ref{sec:execution_optimization}, MAS4TS composes multiple tools into chains where each tool's output serves as input to the next. A tool chain is formally defined as:
\begin{equation}
\hat{\mathbf{Y}} = \mathcal{T}_{m_K} \circ \mathcal{T}_{m_{K-1}} \circ \cdots \circ \mathcal{T}_{m_1}(\mathbf{Z}),
\end{equation}
where $\circ$ denotes functional composition and $\{m_1, \ldots, m_K\}$ are tool indices selected by routing policy $\pi_{\Theta}$.

\paragraph{I/O schema and validity.}
Each tool $\mathcal{T}_m$ exposes (i) an input schema (tensor shape, mask fields, and optional auxiliary context) and (ii) an output schema compatible with the next tool in the chain.
The tool registry enforces schema consistency at composition time; invalid compositions are excluded from $\mathcal{C}_\tau$ (Section~\ref{appx:unified_interface}).
This design makes tool chains auditable and prevents silent shape/type mismatches during routing.

\paragraph{Default Chains.}
Table~\ref{tab:default_chains} shows the default tool chains when routing confidence exceeds 0.6. Brackets indicate conditionally included components based on priors $\mathcal{P}$.

\begin{table}[h]
\centering
\caption{Default Tool Chains by Task}
\label{tab:default_chains}
\begin{tabular}{ll}
\toprule
\textbf{Task} & \textbf{Tool Chain} \\
\midrule
Forecasting & RevIN $\rightarrow$ [Decomposition] $\rightarrow$ PatchTransformer $\rightarrow$ RevIN$^{-1}$ \\
Classification & Normalization $\rightarrow$ TimesBlock $\rightarrow$ GlobalPool $\rightarrow$ ClassificationHead \\
Imputation & MaskEncode $\rightarrow$ Normalization $\rightarrow$ PatchTransformer $\rightarrow$ Reconstruction \\
Anomaly Detection & InstanceNorm $\rightarrow$ MultiScaleEncoder $\rightarrow$ Reconstruction $\rightarrow$ ScoreCompute \\
\bottomrule
\end{tabular}
\end{table}

\paragraph{Ensemble Chaining.}
When routing confidence is low ($< 0.6$), multiple tool chains can be ensembled:
\begin{equation}
\hat{\mathbf{Y}} = \sum_{i=1}^{K} w_i \cdot \text{Chain}_i(\mathbf{Z}), \quad w_i = \frac{\exp(s_i)}{\sum_j \exp(s_j)},
\end{equation}
where $s_i$ is the confidence score computed by $g_{\psi}$ for chain $i$. This provides robustness under ambiguous or out-of-distribution inputs.

\paragraph{Practical note on determinism.}
To reduce variance during evaluation, we disable stochastic sampling over chains and use deterministic top-1 selection when $c \ge 0.6$, and deterministic top-$k$ ensembling otherwise (Section~\ref{appx:unified_interface}).
This makes routing behavior stable across runs, given a fixed random seed.

\newpage

%% file: appendix/prompts.tex
\section{VLM Prompt Templates}
\label{appx:prompt}

This appendix documents the prompt templates used by the \textbf{Reasoner} agent for visual reasoning, as introduced in Section~\ref{subsec:visual_reasoning}. Each prompt is designed to elicit structured, task-specific outputs from Vision-Language Models that can be directly consumed by downstream numerical reasoning.

\subsection{Prompt Design Principles}
\label{appx:prompt_philosophy}

Our prompts follow three key principles:
\begin{enumerate}[leftmargin=*, itemsep=2pt]
    \item \textbf{Structured Output}: All prompts request JSON-formatted responses with predefined schema, ensuring deterministic parsing and seamless integration with downstream modules.
    \item \textbf{Statistical Grounding}: Input statistics (min, max, mean, std) and semantic context are provided to anchor visual reasoning in quantitative reality, reducing out-of-range hallucinations.
    \item \textbf{Task-Specific Anchors}: Different tasks require different anchor types---forecasting anchors define future trajectory shape, anomaly anchors contrast normal vs.\ abnormal patterns, classification anchors capture discriminative signatures, and imputation anchors guide local reconstruction.
\end{enumerate}

The anchor count range \texttt{\{min\_anchors\}-\{max\_anchors\}} is user-configurable. We empirically suggest: forecasting (8--15), anomaly detection (8--12), classification (5--8), and imputation (5--7), where longer prediction horizons generally benefit from denser anchors.

%==============================================================================
\subsection{Forecasting Prompt}
\label{appx:prompt_forecast}

\begin{tcolorbox}[colback=gray!5, colframe=gray!50, title=System Prompt (Forecasting)]
\small
\texttt{You are a senior time-series forecasting expert. Analyze the time series plot and provide predictions in strict JSON format. Always include a confidence score (0.00-1.00) based on pattern consistency.}
\end{tcolorbox}

\begin{tcolorbox}[colback=blue!5, colframe=blue!30, title=User Prompt Template (Forecasting)]
\small
\texttt{Analyze this time series plot for FORECASTING task (predict \{pred\_len\} future steps).}

\texttt{Input sequence: \{seq\_len\} steps (t=0 to \{seq\_len-1\}), last value=\{last\_value\}}

\texttt{Historical stats: range=[\{history\_min\}, \{history\_max\}], std=\{history\_std\}}

\texttt{Requirements:}
\begin{enumerate}[leftmargin=*, itemsep=0pt]
    \item \texttt{Output ONLY \{min\_anchors\}-\{max\_anchors\} key anchor points (peaks/valleys/inflections) for the PREDICTION WINDOW}
    \item \texttt{Anchor types: 'start', 'peak', 'valley', 'inflection', 'end'}
    \item \texttt{All values MUST stay within reasonable bounds}
    \item \texttt{Confidence score reflects pattern continuity}
\end{enumerate}

\texttt{Output JSON:}
\begin{verbatim}
{
  "confidence": 0.85,
  "anchors": [
    {"t": 96, "v": 0.342, "type": "start"},
    {"t": 120, "v": 0.456, "type": "peak"},
    ...
  ]
}
\end{verbatim}
\end{tcolorbox}

The forecasting prompt explicitly constrains anchors to the prediction window (timesteps $t \geq L$) rather than summarizing historical patterns. The \texttt{last\_value} field ensures continuity at the forecast boundary, while \texttt{history\_std} helps calibrate reasonable fluctuation magnitudes. These anchors are subsequently fed to the Latent ODE decoder for dense trajectory reconstruction.

%==============================================================================
\subsection{Classification Prompt}
\label{appx:prompt_classification}

\begin{tcolorbox}[colback=gray!5, colframe=gray!50, title=System Prompt (Classification)]
\small
\texttt{You are a senior time-series classification expert. Analyze the time series plot and classify the sequence pattern. Provide strict JSON with confidence score.}
\end{tcolorbox}

\begin{tcolorbox}[colback=green!5, colframe=green!70, title=User Prompt Template (Classification)]
\small
\texttt{Analyze this time series plot for CLASSIFICATION (input length=\{seq\_len\}).}

\texttt{Input stats: range=[\{history\_min\}, \{history\_max\}], mean=\{history\_mean\}}

\texttt{Requirements:}
\begin{enumerate}[leftmargin=*, itemsep=0pt]
    \item \texttt{Identify the dominant pattern type (periodic, trending, stationary, etc.)}
    \item \texttt{Key anchors: \{min\_anchors\}-\{max\_anchors\} points that define the class signature}
    \item \texttt{Confidence score reflects pattern uniqueness}
\end{enumerate}

\texttt{Output JSON:}
\begin{verbatim}
{
  "confidence": 0.78,
  "pattern_type": "periodic",
  "key_anchors": [
    {"t": 12, "v": 0.89, "type": "period_peak"},
    {"t": 36, "v": -0.45, "type": "period_valley"}
  ]
}
\end{verbatim}
\end{tcolorbox}

The prompt focuses on discriminative temporal features that characterize different classes. 
%Classification anchors are intentionally sparse, capturing signature patterns (e.g., periodic peaks/valleys) rather than exhaustive sequence description. 
The \texttt{pattern\_type} field provides high-level semantic guidance that can be fused with learned numerical features.

%==============================================================================
\subsection{Imputation Prompt}
\label{appx:prompt_imputation}

\begin{tcolorbox}[colback=gray!5, colframe=gray!50, title=System Prompt (Imputation)]
\small
\texttt{You are a senior time-series imputation expert. Analyze the time series plot and impute MISSING VALUES in the INPUT sequence. Provide JSON with confidence score.}
\end{tcolorbox}

\begin{tcolorbox}[colback=yellow!5, colframe=yellow!90, title=User Prompt Template (Imputation)]
\small
\texttt{Analyze this time series plot for IMPUTATION (input length=\{seq\_len\}).}

\texttt{Missing values: marked at positions \{missing\_positions\}}

\texttt{Input stats: range=[\{history\_min\}, \{history\_max\}], mean=\{history\_mean\}}

\texttt{Requirements:}
\begin{enumerate}[leftmargin=*, itemsep=0pt]
    \item \texttt{Impute ONLY marked missing positions}
    \item \texttt{Specify reasoning: 'interpolation' or 'extrapolation'}
    \item \texttt{Key anchors: \{min\_anchors\}-\{max\_anchors\} critical points guiding imputation}
\end{enumerate}

\texttt{Output JSON:}
\begin{verbatim}
{
  "confidence": 0.88,
  "imputed_values": [
    {"t": 25, "v": 0.34, "reason": "interpolation"},
  ],
  "key_anchors": [
    {"t": 20, "v": 0.28, "type": "observed"},
  ]
}
\end{verbatim}
\end{tcolorbox}

The imputation prompt explicitly marks missing positions to focus the VLM's attention on relevant gaps. The \texttt{reason} field distinguishes interpolation (bounded by observations on both sides) from extrapolation (extending from one-sided context), informing the Latent ODE about reconstruction confidence. Key anchors at observed boundaries provide hard constraints that ensure consistency with known data points.

\subsection{Anomaly Detection Prompt}
\label{appx:prompt_anomaly}

\begin{tcolorbox}[colback=gray!5, colframe=gray!50, title=System Prompt (Anomaly Detection)]
\small
\texttt{You are a senior time-series anomaly detection expert. Analyze the time series plot and identify anomalous time steps in the INPUT sequence. Provide strict JSON with confidence scores.}
\end{tcolorbox}

\begin{tcolorbox}[colback=red!5, colframe=red!30, title=User Prompt Template (Anomaly Detection)]
\small
\texttt{Analyze this time series plot for ANOMALY DETECTION (input length=\{seq\_len\}).}

\texttt{Input stats: range=[\{history\_min\}, \{history\_max\}], mean=\{history\_mean\}, std=\{history\_std\}}

\texttt{Requirements:}
\begin{enumerate}[leftmargin=*, itemsep=0pt]
    \item \texttt{Focus ONLY on INPUT SEQUENCE (t=0 to \{seq\_len-1\})}
    \item \texttt{Anomaly scores: 0.0-1.0 (higher = more anomalous)}
    \item \texttt{For consecutive anomalies, report ONLY the most significant one}
    \item \texttt{Key anchors: \{min\_anchors\}-\{max\_anchors\} points including normal patterns as reference}
\end{enumerate}

\texttt{Output JSON:}
\begin{verbatim}
{
  "confidence": 0.82,
  "anomaly_scores": [
    {"t": 45, "score": 0.91, "reason": "spike"},
    {"t": 78, "score": 0.85, "reason": "level_shift"}
  ],
  "key_anchors": [
    {"t": 10, "v": 0.23, "type": "normal"},
    {"t": 45, "v": 1.82, "type": "anomaly"}
  ]
}
\end{verbatim}
\end{tcolorbox}

The anomaly detection prompt requests both anomaly scores and reference anchors. This dual-output structure enables the reconstruction-based detection pipeline to distinguish anomalous patterns from normal temporal dynamics. The \texttt{reason} field (spike, level\_shift, etc.) provides interpretability for downstream analysis. Requiring only the most significant anomaly among consecutive ones prevents redundant detections.

%==============================================================================

\newpage

%% file: appendix/justification.tex
\section{Justification of MAS}
Although forecasting, imputation, anomaly detection, and classification differ in outputs and objectives, they share a common workflow: (i) \textbf{perceive} salient temporal structures, (ii) \textbf{reason} about global shape constraints and local irregularities, and (iii) \textbf{execute} decisions under validity constraints (e.g., consistency with observations, boundedness, and smoothness). This aligns naturally with an agentic view and motivates modeling time-series analysis as analysis--reasoning--execution pipeline.

A single monolithic model typically entangles these steps into one latent representation with a fixed head, which makes it hard to (a) adapt the execution pathway to heterogeneous series characteristics and task requirements, and (b) incorporate domain tools (e.g., decomposition, robust statistics, constraint projection) in an auditable and controllable manner. A multi-agent design provides a principled way to modularize the workflow into specialized roles while still enabling coordination via a controlled shared-memory interface. Specifically, MAS4TS follows an \textbf{Analyzer--Reasoner--Executor (ARE)} paradigm: the \emph{Analyzer} produces grounded statistical priors and a plot-based interface, the \emph{Reasoner} converts perceived structures into explicit anchors and trajectory constraints, and the \emph{Executor} selects, composes, and verifies task-specific tool chains.

This separation supports adaptive computation (different tool compositions for different regimes), improves interpretability (anchors, constraints, and verifier checks are inspectable), and reduces training burden by focusing learning on coordination-critical components (e.g., routing and memory gating) rather than retraining task-specific backbones. In this sense, the benefit of MAS is not merely parallelization, but the ability to unify diverse time-series tasks via a shared reasoning-and-execution protocol with explicit intermediate artifacts.

%% file: appendix/limitation.tex
\section{Limitations and Future Work}
MAS4TS inherits the capabilities and failure modes of its underlying foundation models (especially the VLM used for anchor extraction), which can affect robustness under distribution shift. In addition, for complex multivariate and long-context scenarios, the agent workflow may introduce redundant context and communication overhead, potentially reducing the efficiency gains from tool-driven execution.

We will continue to extend MAS4TS as an evolving time-series agent system by (i) enriching the tool library and improving tool routing for harder real-world constraints, and (ii) exploring agent \textbf{skills} and foundation-model reasoning strategies to better handle complex contexts with less redundancy and stronger generalization.